\documentclass[10pt,journal,compsoc]{IEEEtran}
\usepackage{amsmath,amsfonts}
\usepackage{cite}
\usepackage{array}
\usepackage{booktabs}
\usepackage{graphicx}
\usepackage{textcomp}
\usepackage{stfloats}
\usepackage{url}
\usepackage{xcolor}
\usepackage{tikz}
\usetikzlibrary{arrows.meta,patterns}
\usepackage[hidelinks]{hyperref}
\usepackage{xurl}
\begin{document}
\bstctlcite{IEEEexample:BSTcontrol}

\title{Machine Learning for Live-Birth Prediction After IVF/ICSI From Structured Clinical Data: A Systematic Review, Meta-Analysis, and Critical Appraisal}

\author{Abdul~Raheem~Shahzad, Muhammad~Zaheer~Khan, Umar~Ajaib, Asad~Ur~Rehman, Juan~Deng, and~Ming~Zhang%
\IEEEcompsocitemizethanks{\IEEEcompsocthanksitem A.R. Shahzad, M.Z. Khan, U. Ajaib, A.U. Rehman and J. Deng are with the School of Computer Science, Wuhan University, Wuhan 430072, Hubei, China.\protect\\
E-mail: abdul.raheem@whu.edu.cn, zaheer@whu.edu.cn, umar.ajaib@whu.edu.cn, asadurrehman@whu.edu.cn, dengjuan@whu.edu.cn
\IEEEcompsocthanksitem M. Zhang is with the Reproductive Medical Center, Zhongnan Hospital of Wuhan University, and the Clinical Medicine Research Center of Prenatal Diagnosis and Birth Health in Hubei Province, Wuhan, Hubei, China.\protect\\
E-mail: whu\_mz@163.com
\IEEEcompsocthanksitem Corresponding author: Abdul Raheem Shahzad.\protect\\
This work has been submitted to the IEEE for possible publication. Copyright may be transferred without notice, after which this version may no longer be accessible. Data, code and the supplementary material: \url{https://github.com/abdul-raheem-shahzad/ml-ivf-live-birth-review}.}}

\markboth{Preprint submitted to IEEE Transactions on Pattern Analysis and Machine Intelligence}%
{Shahzad \MakeLowercase{\textit{et al.}}: Machine Learning for Live-Birth Prediction After IVF/ICSI}

\IEEEtitleabstractindextext{%
\begin{abstract}
Machine-learning models predicting live birth after in vitro fertilisation (IVF) or intracytoplasmic sperm injection (ICSI) from routine clinical data report performance from near chance to near perfect. We systematically reviewed these models under PRISMA 2020, searching OpenAlex, Scopus and Web of Science to 22 October 2026, appraised them with PROBAST+AI and TRIPOD+AI, and pooled their discrimination with random-effects meta-analysis on the logit scale. We define what fixes the meaning of a model's output (decision point, endpoint, unit of analysis, data source) and map the 48 included studies by model family, decision point and validation design. Models evaluated only internally pooled to an area under the receiver operating characteristic curve (AUC) of 0.814, against 0.756 (95\% prediction interval 0.546--0.889) for models evaluated on a later period or another centre, and 0.764 for studies not at high risk of bias; neither difference was significant, and heterogeneity was extreme. Models whose decision point was not stated pooled to 0.910, the only significant moderator in meta-regression. Against logistic regression fitted to the same data, the best machine-learning model gained a pooled 0.056 AUC (95\% CI 0.017--0.090). Of 48 studies, 47 were at high or unclear risk of bias, only four assessed the calibration slope, and no model has evidence of clinical impact. The certainty of the pooled evidence is low to very low; the gains that survive validation came from staging predictions to decision points and local calibration, not deeper architectures. We provide a TRIPOD+AI-mapped checklist, a benchmark specification and a research agenda
\end{abstract}

}

\maketitle
\IEEEdisplaynontitleabstractindextext
\IEEEpeerreviewmaketitle

\section{Introduction}
\IEEEPARstart{I}{nfertility} affects about one in six people during their lifetime \cite{cox2022hropen}. In vitro fertilisation (IVF) and intracytoplasmic sperm injection (ICSI) are the main treatments, yet the clinical pregnancy rate per cycle is about 30--40\%, and the live-birth rate is lower \cite{gao2025patterns}. Each cycle carries physical, emotional and financial costs, so patients and clinicians need an individual estimate of the chance of a live birth: before treatment, to decide whether to start and how to fund it, and during treatment, to choose between fresh transfer and freeze-all and how many embryos to transfer \cite{mclernon2023bpr,eshre2024embryos}.

Regression models have served these purposes for three decades. Templeton et~al.~\cite{templeton1996lancet} derived the first widely used model from UK registry data, IVFpredict \cite{nelson2011plosmed} and the Society for Assisted Reproductive Technology (SART) model \cite{luke2014fs} followed, and the discrete-time models of McLernon et~al.~\cite{mclernon2016bmj}, fitted to 113,873 UK women, remain the reference \cite{tian2025humrep}. Machine learning entered early: Banerjee et~al.~\cite{banerjee2010pnas} fitted boosted trees to 52 clinical and embryological variables and validated them on a later period. The field has since grown quickly \cite{gao2025patterns,ouyang2026abme}, and reported areas under the receiver operating characteristic curve (AUC) for live birth now range from near chance to near perfect.

Two literatures have developed in parallel (Table~\ref{tbl:priorreviews}). Reviews of artificial intelligence (AI) in IVF concentrate on embryo assessment \cite{ouyang2026abme}, and the one systematic review of AI live-birth prediction pooled image-based and clinical models and appraised them with a diagnostic-accuracy tool, without meta-analysis \cite{yadav2026mefsj}. Reviews of IVF prediction models appraise methodology rigorously but focus on regression and cumulative outcomes \cite{leushuis2009hru,mclernon2023bpr,tian2025humrep}. Meanwhile, the standards for prediction models have been updated for machine learning in TRIPOD+AI and PROBAST+AI \cite{collins2024tripodai,moons2025probastai}, and across medicine the comparisons that favour machine learning over logistic regression are those at high risk of bias \cite{christodoulou2019jce}. No review combines a systematic search, an appraisal against these AI-specific standards and a quantitative synthesis stratified by validation design for the task most clinics can address with data they already hold: predicting live birth from structured clinical data.

\begin{table*}[!t]
\caption{Previous reviews of prediction models for IVF outcomes and how this review differs.}\label{tbl:priorreviews}
\scriptsize
\setlength{\tabcolsep}{3.5pt}
\renewcommand{\arraystretch}{1.12}
\begin{tabular}{@{}>{\raggedright\arraybackslash}p{\dimexpr 0.1295\textwidth - 2\tabcolsep - 0.143pt\relax}>{\raggedright\arraybackslash}p{\dimexpr 0.1295\textwidth - 2\tabcolsep - 0.143pt\relax}>{\raggedright\arraybackslash}p{\dimexpr 0.2073\textwidth - 2\tabcolsep - 0.143pt\relax}>{\raggedright\arraybackslash}p{\dimexpr 0.1140\textwidth - 2\tabcolsep - 0.143pt\relax}>{\raggedright\arraybackslash}p{\dimexpr 0.1140\textwidth - 2\tabcolsep - 0.143pt\relax}>{\raggedright\arraybackslash}p{\dimexpr 0.1399\textwidth - 2\tabcolsep - 0.143pt\relax}>{\raggedright\arraybackslash}p{\dimexpr 0.1658\textwidth - 2\tabcolsep - 0.143pt\relax}@{}}
\toprule
\textbf{Review} & \textbf{Design} & \textbf{Scope} & \textbf{Outcome} & \textbf{Studies} & \textbf{Appraisal} & \textbf{Synthesis} \\
\midrule
Leushuis et~al.~\cite{leushuis2009hru} & Systematic review & Prediction models in reproductive medicine, mainly regression & Pregnancy outcomes & -- & Derivation, validation and impact phases & Narrative \\
McLernon and Bhattacharya~\cite{mclernon2023bpr} & Narrative review & Clinical prediction models in IVF; important covariates & Cumulative live birth & -- & Model quality, qualitative & Narrative \\
Tian et~al.~\cite{tian2025humrep} & Systematic review, meta-analysis and external validation & Clinical prediction models for IVF outcomes, regression and machine learning & Live birth and other IVF outcomes & 72 studies; 86 live-birth models & Risk-of-bias assessment & Pooled discrimination; external validation in a Chinese cohort \\
Gao et~al.~\cite{gao2025patterns} & Narrative review & AI across the IVF pathway by data modality & Several & -- & None formal & Narrative \\
Ouyang and Wei~\cite{ouyang2026abme} & Review & Multimodal AI for embryo grading and pregnancy prediction & Pregnancy (various definitions) & 17 tabular and 8 multimodal pregnancy studies tabulated & None formal & Tabulated AUCs \\
Yadav et~al.~\cite{yadav2026mefsj} & Systematic review without meta-analysis & AI live-birth prediction, image-based and clinical models & Live birth & 23 studies & QUADAS-2 & Narrative \\
This review & Systematic review, meta-analysis and critical appraisal & Machine-learning models of live birth from structured clinical data, with regression benchmarks & Live birth per transfer, per cycle or cumulative & 48 studies; 66 AUC estimates & PROBAST+AI; TRIPOD+AI & Random-effects meta-analysis of logit AUC by validation design, decision point and risk of bias; paired comparisons with logistic regression \\
\bottomrule
\end{tabular}

\vspace{3pt}
{\scriptsize -- Not applicable or not reported. QUADAS-2 is a tool for diagnostic-accuracy studies; PROBAST+AI and TRIPOD+AI are designed for prediction models.\par}
\end{table*}

This review asks four questions. RQ1: which design choices determine what the output of a live-birth model means? RQ2: how do model families perform, and how does reported discrimination depend on validation design, decision point and risk of bias? RQ3: how do published models fare against PROBAST+AI and TRIPOD+AI? RQ4: what is needed for clinical translation, and which emerging methods are worth pursuing? Its contributions are a PRISMA 2020 systematic search with a full screening log \cite{page2021bmj}; a decision-point framework that fixes the admissible predictors for each clinical question; a taxonomy that crosses model family, decision point and validation design; a random-effects meta-analysis of AUC stratified by validation design, decision point and risk of bias, with paired comparisons against logistic regression; a PROBAST+AI appraisal of every included study; and a 16-item checklist mapped to TRIPOD+AI, with a benchmark specification and a research agenda.

Section~2 describes the methods, Section~3 the clinical and data context (RQ1), Section~4 model families and pooled performance (RQ2), and Section~5 methodological quality (RQ3). Sections~6 and 7 address translation and emerging methods (RQ4), Section~8 gives recommendations, and Sections~9 and 10 state the limitations and conclusions. Per-study tables, extended discussion and the search strings are in Supplementary Appendices A--G. Three of the authors have developed a live-birth model available only as a preprint \cite{shahzad2026arxiv}; under the preprint rule of Section~2.1 it is excluded from the evidence tables, the appraisal and the meta-analysis.

\section{Review Methods}
The review was conducted and reported according to PRISMA 2020 \cite{page2021bmj} (completed checklist: Supplementary File S6). Data extraction followed CHARMS \cite{moons2014charms}, and risk of bias was appraised with PROBAST+AI \cite{moons2025probastai}. The protocol was not registered.

\subsection{Eligibility Criteria and Information Sources}
Primary studies were eligible if they developed or validated a multivariable model that predicts live birth or cumulative live birth after IVF or ICSI from structured clinical, laboratory or embryological data, fitted at least one learner other than logistic or penalised logistic regression, and reported discrimination for live birth as an AUC or c-statistic. Fresh, frozen, donor-oocyte and preimplantation-genetic-testing (PGT) cycles were eligible, at any decision point and with live birth defined per transfer, per cycle or cumulatively. Embryo grades recorded as variables were allowed; models whose inputs were images, videos, omics or microbiome data were excluded unless a structured-data model was reported separately. We excluded studies of clinical pregnancy or implantation only, studies pooling IVF with intrauterine insemination, reviews, protocols, editorials and conference abstracts. Preprints were excluded unless a peer-reviewed version existed. Regression-only models \cite{templeton1996lancet,nelson2011plosmed,luke2014fs,mclernon2016bmj} are discussed as benchmarks but are not part of the machine-learning evidence set.

We searched three databases without date, language or document-type limits (Supplementary Appendix A). OpenAlex, which aggregates PubMed/MEDLINE, Crossref, repositories and preprint servers \cite{priem2022openalex}, was searched on 21 October 2026 with four title-and-abstract queries and on 22 October 2026 with five supplementary queries that added deep learning, artificial intelligence, XGBoost, gradient boosting and support vector machines. Scopus and the Web of Science Core Collection were searched on 22 October 2026 with one query each that combined live birth, IVF or ICSI and twelve machine-learning terms. Reference lists of the included studies and of the reviews in Table~\ref{tbl:priorreviews} were screened.

\subsection{Study Selection, Data Extraction and Appraisal}
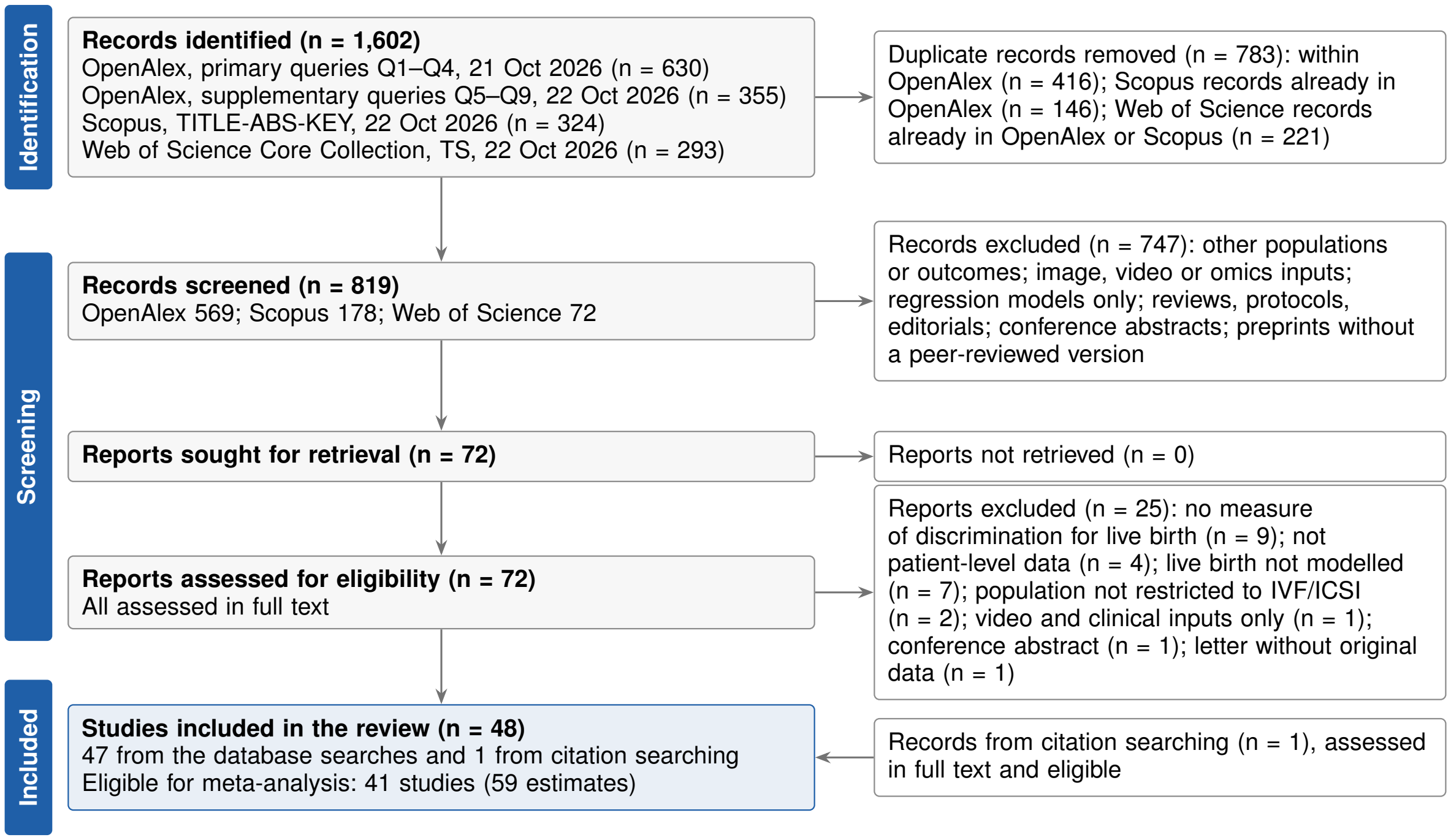
\begin{figure*}[!t]
\centering
\resizebox{0.86\textwidth}{!}{
\begin{tikzpicture}[every node/.append style={execute at begin node={\hyphenpenalty=10000\relax\exhyphenpenalty=10000\relax}},x=1cm,y=1cm,font=\sffamily\scriptsize,
  box/.style={draw=black!45,fill=black!3,rounded corners=1.5pt,align=left,inner sep=4pt,text width=#1},
  box/.default=7.4cm,
  side/.style={draw=black!45,fill=white,rounded corners=1.5pt,align=left,inner sep=4pt,text width=5.6cm},
  stage/.style={fill=stageblue,text=white,font=\sffamily\scriptsize\bfseries,rounded corners=1.5pt,minimum height=1.0cm,text width=0.42cm,align=center,inner sep=1pt},
  arr/.style={-{Stealth[length=1.6mm]},draw=black!55,line width=0.55pt}]
\definecolor{stageblue}{HTML}{1F5FA8}
\node[box] (id) at (0,0.75) {\textbf{Records identified (n = 1,602)}\\
OpenAlex, primary queries Q1--Q4, 21 Oct 2026 (n = 630)\\
OpenAlex, supplementary queries Q5--Q9, 22 Oct 2026 (n = 355)\\
Scopus, TITLE-ABS-KEY, 22 Oct 2026 (n = 324)\\
Web of Science Core Collection, TS, 22 Oct 2026 (n = 293)};
\node[side,anchor=west] (dup) at (4.45,0.75) {Duplicate records removed (n = 783): within OpenAlex (n = 416); Scopus records already in OpenAlex (n = 146); Web of Science records already in OpenAlex or Scopus (n = 221)};
\node[box] (scr) at (0,-1.35) {\textbf{Records screened (n = 819)}\\
OpenAlex 569; Scopus 178; Web of Science 72};
\node[side,anchor=west] (scrx) at (4.45,-1.35) {Records excluded (n = 747): other populations or outcomes; image, video or omics inputs; regression models only; reviews, protocols, editorials; conference abstracts; preprints without a peer-reviewed version};
\node[box] (sought) at (0,-2.95) {\textbf{Reports sought for retrieval (n = 72)}};
\node[side,anchor=west] (nr) at (4.45,-2.95) {Reports not retrieved (n = 0)};
\node[box] (elig) at (0,-4.35) {\textbf{Reports assessed for eligibility (n = 72)}\\
All assessed in full text};
\node[side,anchor=west] (eligx) at (4.45,-4.35) {Reports excluded (n = 25): no measure of discrimination for live birth (n = 9); not patient-level data (n = 4); live birth not modelled (n = 7); population not restricted to IVF/ICSI (n = 2); video and clinical inputs only (n = 1); conference abstract (n = 1); letter without original data (n = 1)};
\node[box,fill=stageblue!10,draw=stageblue] (inc) at (0,-6.05) {\textbf{Studies included in the review (n = 48)}\\
47 from the database searches and 1 from citation searching\\
Eligible for meta-analysis: 41 studies (59 estimates)};
\node[side,anchor=west] (oth) at (4.45,-6.05) {Records from citation searching (n = 1), assessed in full text and eligible};
\draw[arr] (id) -- (scr); \draw[arr] (scr) -- (sought); \draw[arr] (sought) -- (elig); \draw[arr] (elig) -- (inc);
\draw[arr] (oth.west) -- (inc.east);
\draw[arr] (id.east) -- (dup.west); \draw[arr] (scr.east) -- (scrx.west); \draw[arr] (sought.east) -- (nr.west); \draw[arr] (elig.east) -- (eligx.west);
\node[stage,minimum height=1.9cm] at (-4.25,0.75) {\rotatebox{90}{Identification}};
\node[stage,minimum height=4.0cm] at (-4.25,-2.85) {\rotatebox{90}{Screening}};
\node[stage,minimum height=1.6cm] at (-4.25,-6.05) {\rotatebox{90}{Included}};
\end{tikzpicture}}
\caption{PRISMA 2020 flow diagram of the search and study selection. The screening log for all 819 screened records, with their source database, is given in Supplementary File S1.}\label{fig:prisma}
\end{figure*}

Records were deduplicated on DOI, or on title where no DOI existed, screened on title and abstract, and assessed in full text by the first author (Fig.~\ref{fig:prisma}). The searches identified 1,602 records, of which 819 remained after removing 783 duplicates. All 72 reports sought were obtained in full text and assessed; reports in Chinese, Russian and Spanish were read with machine translation. Twenty-five were excluded: nine reported no measure of discrimination for live birth, four did not use patient-level data, seven modelled other outcomes, two did not restrict the population to IVF or ICSI, and three were excluded for other reasons (video inputs only, conference abstract, letter). Citation searching added one study \cite{banerjee2010pnas}, giving 48 included studies. Scopus added six studies that OpenAlex had missed \cite{benchaib2022rmb,lantzi2025biology,priya2026jitm,liang2021cmpb,baij2026cjrc,xing2025cjrc}; Web of Science added none. Supplementary File S1 lists every record with its decision.

For each study we extracted the setting, data source and period, sample size, events, case mix, cycle type, decision point, endpoint, model families, validation design, discrimination with its uncertainty, the AUC of logistic regression on the same data, calibration, decision-curve analysis, code availability and modelling steps. AI assistance was used to locate reported figures; every value was checked against its source, and two co-authors (M.Z.K., U.A.) verified the extraction, one independently. Risk of bias was rated low, high or unclear in the four PROBAST+AI domains (participants and data sources, predictors, outcome, analysis), with the overall rating low only when every domain was low and high when any domain was high \cite{moons2025probastai}. The first author appraised every study first; two co-authors (M.Z.K., U.A.) then re-appraised all 48 independently of each other and of the first author, with the same written protocol, without access to the manuscript, the supplement or the first author's ratings; each read the full texts and used an AI model (Claude, Anthropic) to draft and check the ratings. Agreement was measured by raw agreement, Cohen's kappa and Gwet's AC1. Disagreements were adjudicated by re-reading the source. Ratings, evidence and every change are in Supplementary Tables S3, S7--S8 and E2; extracted data in Table S2.

\subsection{Quantitative Synthesis}
Discrimination was pooled on the logit scale, which brings the between-study distribution of the c-statistic closer to normal \cite{snell2018smmr}. For study $i$ with $\mathrm{AUC}_i$ and 95\% confidence limits $L_i$ and $U_i$,
\begin{equation}
\theta_i=\ln\frac{\mathrm{AUC}_i}{1-\mathrm{AUC}_i},\qquad
\widehat{\mathrm{SE}}(\theta_i)=\frac{\operatorname{logit}(U_i)-\operatorname{logit}(L_i)}{2\times 1.96}.
\label{eq:logit}
\end{equation}
When no interval was reported, the standard error of the AUC $A$ was approximated from the numbers of events $n_1$ and non-events $n_0$ in the evaluation set \cite{hanley1982radiology} and transformed with the delta method:
\begin{equation}
\begin{aligned}
\mathrm{SE}(A)^2&=\bigl[A(1-A)+(n_1-1)(Q_1-A^2)\\
&\qquad+(n_0-1)(Q_2-A^2)\bigr]\big/(n_1 n_0),\\
Q_1&=\frac{A}{2-A},\quad Q_2=\frac{2A^2}{1+A},\quad \mathrm{SE}(\theta)=\frac{\mathrm{SE}(A)}{A(1-A)}.
\end{aligned}
\label{eq:hm}
\end{equation}
Where only the event rate was reported, counts were derived from it; where only the size of the evaluation set was reported, a rate of 40\% was assumed. Estimates without an estimable variance are shown but not pooled, and all approximations are flagged in Supplementary Table S4. Estimates were combined in a random-effects model with the between-study variance estimated by restricted maximum likelihood and Hartung--Knapp--Sidik--Jonkman intervals with the ad hoc variance correction, as recommended for the c-statistic \cite{debray2017bmj,inthout2014bmc}; heterogeneity is reported as $I^2$ \cite{higgins2002sim}, and the 95\% prediction interval uses $k-2$ degrees of freedom \cite{higgins2009jrssa}:
\begin{align}
\hat\tau^2&=\max\Bigl\{0,\tfrac{\sum_i w_i^{2}[(\theta_i-\hat\mu)^2-v_i]}{\sum_i w_i^{2}}+\tfrac{1}{\sum_i w_i}\Bigr\},\ w_i=\tfrac{1}{v_i+\hat\tau^2},\nonumber\\
\hat\mu&=\tfrac{\sum_i w_i\theta_i}{\sum_i w_i},\quad \widehat{\mathrm{SE}}(\hat\mu)^2=\tfrac{\max\{1,q\}}{\sum_i w_i},\ q=\tfrac{\sum_i w_i(\theta_i-\hat\mu)^2}{k-1},\label{eq:re}\\
\mathrm{PI}_{95\%}&=\hat\mu\pm t_{k-2,\,0.975}\sqrt{\hat\tau^2+\widehat{\mathrm{SE}}(\hat\mu)^2},\label{eq:pi}
\end{align}
with the confidence interval $\hat\mu\pm t_{k-1,0.975}\widehat{\mathrm{SE}}(\hat\mu)$ and $v_i$ the within-study variance. Pooled values were back-transformed, strata were compared with a $t$ test on the pooled logits, and DerSimonian--Laird estimation \cite{dersimonian1986cct} was a sensitivity analysis.

Each study contributed at most one estimate per analysis. The primary analysis compared internal validation (random split or cross-validation) with temporal or geographic validation, using each study's estimate at its earliest decision point. Secondary analyses stratified by overall risk of bias and by decision point, using the external estimate where one existed. Models using information recorded after a positive pregnancy test were shown but not pooled. Planned sensitivity analyses used the latest decision point, excluded evaluation sets under 100 patients, assumed event rates, rebalanced or augmented evaluations or models without a stated decision point, and kept one study per group of overlapping cohorts \cite{fu2026fendo,liul2026fendo,liu2024rbejama,cai2024humrep,liuz2024jarg}; exploratory ones kept only reported confidence intervals, separated cumulative from per-cycle endpoints and omitted each study in turn (all in Supplementary Table S10). Where a study reported internal and external estimates at one decision point, the within-study change was pooled, with the variances added because the evaluation sets do not overlap. The moderators were fitted jointly in a meta-regression of logit AUC on external validation, high risk of bias and an unstated decision point, with Knapp--Hartung standard errors \cite{knapp2003statmed} checked by 1,000 permutations \cite{higgins2004sim}, and again with model family added. Because the standard error of a logit AUC depends on the AUC, small-study effects were tested by regressing logit AUC on the inverse square root of the effective sample size rather than with Egger's test alone \cite{egger1997bmj,peters2006jama,debray2018rsm}. Where a study fitted logistic regression and machine learning to the same data, the paired difference was pooled with an assumed within-study correlation of 0.5 (0.25--0.75 in sensitivity analyses). Calibration was too rarely reported to be pooled. The stratified analyses and the planned sensitivity analyses were specified with the review; the others were added after the primary results were known and are exploratory, and no adjustment was made for multiple comparisons. Certainty was rated with GRADE for prognostic evidence \cite{iorio2015bmj} (Supplementary Table D2). Code and data are supplied as supplementary material.

\section{Clinical and Data Context}
Four choices made before any model is fitted determine what its output means: the clinical decision it supports, the endpoint and its denominator, the unit of analysis, and the data source.

\subsection{Decision Points in the Treatment Pathway}
\begin{figure*}[!t]
\centering
\resizebox{\textwidth}{!}{
\begin{tikzpicture}[x=1cm,y=1cm,font=\sffamily\scriptsize,
  stage/.style={draw=black!45,fill=black!4,rounded corners=2pt,minimum width=2.25cm,minimum height=1.05cm,align=center,inner sep=2pt},
  dp/.style={circle,fill=dpblue,text=white,font=\sffamily\scriptsize\bfseries,inner sep=1.2pt,minimum size=4.2mm},
  arr/.style={-{Stealth[length=1.8mm]},draw=black!55,line width=0.6pt}]
\definecolor{dpblue}{HTML}{1F5FA8}
\definecolor{cbase}{HTML}{2A78D6}
\definecolor{cstim}{HTML}{1BAF7A}
\definecolor{clab}{HTML}{EB6834}
\foreach \i/\t in {0/{Pre-treatment\\assessment},1/{Ovarian\\stimulation},2/{Oocyte retrieval\\and fertilisation},3/{Embryo\\culture},4/{Embryo\\transfer},5/{Pregnancy\\test},6/{Live birth\\($\geq$ 24 weeks)}}{
  \node[stage] (s\i) at ({1.125+2.55*\i},0) {\t};
}
\foreach \i [evaluate=\i as \j using int(\i+1)] in {0,...,5}{\draw[arr] (s\i.east) -- (s\j.west);}
\foreach \x/\n/\lab in {2.4/1/{Start IVF?\\Counselling, funding},4.95/2/{Trigger; plan fresh\\transfer or freeze-all},10.05/3/{Fresh or freeze-all;\\number of embryos}}{
  \draw[black!35,line width=0.5pt,densely dotted] (\x,0.62) -- (\x,-2.86);
  \node[dp] at (\x,0.98) {\n};
  \node[align=center,font=\sffamily\scriptsize,text=black!80] at (\x,1.62) {\lab};
}
\draw[arr,dpblue,line width=0.7pt] (s5.north) -- ++(0,1.75) -| (s1.north);
\node[dp] at (8.775,2.3) {4};
\node[font=\sffamily\scriptsize,text=black!80,anchor=south west,inner sep=1.5pt] at (9.02,2.31) {After an unsuccessful complete cycle: start another stimulation?};
\def\bh{0.46}
\fill[cbase!22] (0,-1.15) rectangle (17.55,{-1.15-\bh});
\node[anchor=west,text=black!85] at (0.08,{-1.15-\bh/2}) {Baseline: female age, BMI, ovarian reserve (AMH, AFC, FSH), infertility history and cause; male age and semen analysis};
\fill[cstim!22] (2.55,-1.75) rectangle (17.55,{-1.75-\bh});
\node[anchor=west,text=black!85] at (2.63,{-1.75-\bh/2}) {Stimulation: protocol, gonadotropin dose and duration, trigger-day hormones, endometrial thickness};
\fill[clab!22] (5.1,-2.35) rectangle (17.55,{-2.35-\bh});
\node[anchor=west,text=black!85] at (5.18,{-2.35-\bh/2}) {Laboratory and embryology: oocytes, MII, 2PN, embryo development, morphology grade, embryos transferred};
\fill[pattern=north east lines,pattern color=black!35] (12.75,-2.95) rectangle (17.55,{-2.95-\bh});
\draw[black!40,line width=0.4pt] (12.75,-2.95) rectangle (17.55,{-2.95-\bh});
\node[anchor=west,fill=white,inner sep=1pt,text=black!85] at (12.82,{-2.95-\bh/2}) {Post-transfer: hCG, ultrasound};
\node[anchor=east,text=black!70,font=\sffamily\scriptsize,align=right] at (12.6,{-2.95-\bh/2}) {Outcome information: never a predictor at decision points 1\textendash 3};
\node[anchor=west,text=black!60,font=\sffamily\scriptsize] at (0,-3.75) {Shaded bars show when each predictor domain becomes available; a variable is admissible at a decision point only if its bar begins to the left of that point.};
\end{tikzpicture}}
\caption{Decision points in an IVF/ICSI cycle and the availability of predictor domains. Circled numbers mark the four decision points defined in Section~3.1; shaded bars show when each predictor domain becomes known. Post-transfer information is outcome information and is never admissible at decision points 1--3.}\label{fig:pathway}
\end{figure*}
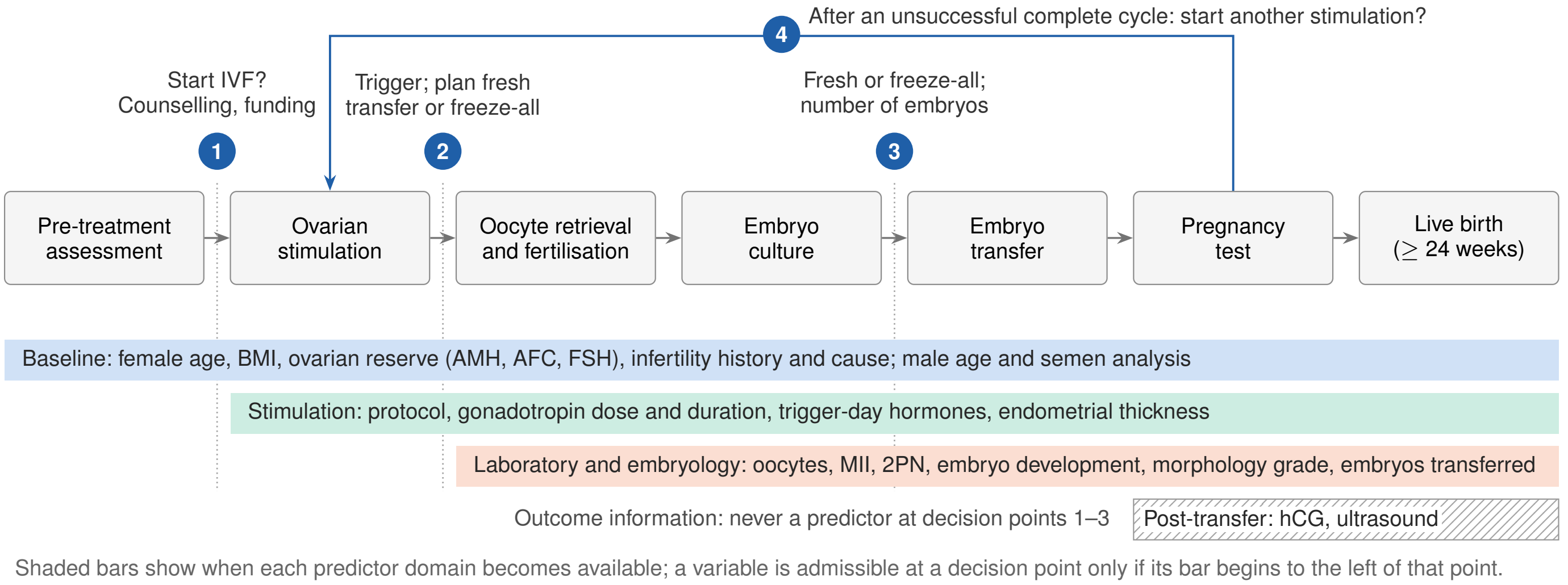

An IVF/ICSI cycle runs through pre-treatment assessment, ovarian stimulation and retrieval, fertilisation, embryo culture, transfer and pregnancy testing, and a prediction is useful only if it informs a decision on this path \cite{gao2025patterns}. The decision fixes which information the model may legitimately use. We distinguish four decision points (Fig.~\ref{fig:pathway}; Table~\ref{tbl:decision}). Before treatment, only baseline information is available (age, body mass index (BMI), ovarian reserve, infertility history and semen analysis), and the questions concern whether to start and what chance to expect; such a model should be used only before the first cycle \cite{mclernon2023bpr}, and Zhang et~al.~\cite{zhang2022frontendo} built one by removing every register field on retrieval, fertilisation, transfer and embryo storage. After stimulation, protocol, dose, trigger-day hormones and endometrial thickness are known, and the decisions concern retrieval and freeze-all. Before transfer, oocyte yield, fertilisation and embryo morphology are known, and the decisions concern fresh transfer and the number and choice of embryos; single embryo transfer is now standard \cite{eshre2024embryos}. After an unsuccessful complete cycle, the decision is whether to start another \cite{banerjee2010pnas,shao2026hro}. A fifth time point, after a positive pregnancy test, supports no treatment decision, and models that use serum $\beta$-hCG after transfer \cite{lin2026fendo,mourad2023fs} or are fitted only to pregnant women \cite{amini2021ijfs,bai2025bmc} give a prognosis, not a prediction comparable with pre-transfer models.

A variable is admissible only if it is routinely known when the decision is made. Guo et~al.~\cite{guo2026frontendo} made this a rule: variables enter cumulatively by stage, and no later-stage variable enters an earlier model. Their best AUC rose from 0.678 at baseline to 0.725 after stimulation and only to 0.731 before transfer, and Liu et~al.~\cite{liul2026fendo} found a similar gradient (0.759 to 0.793). An AUC reported without its decision point cannot be compared with these figures.

\begin{table*}[!t]
\caption{Decision points in an IVF/ICSI cycle and the predictors admissible at each.}\label{tbl:decision}
\scriptsize
\setlength{\tabcolsep}{3.5pt}
\renewcommand{\arraystretch}{1.12}
\begin{tabular}{@{}>{\raggedright\arraybackslash}p{\dimexpr 0.1440\textwidth - 2\tabcolsep - 0.200pt\relax}>{\raggedright\arraybackslash}p{\dimexpr 0.1994\textwidth - 2\tabcolsep - 0.200pt\relax}>{\raggedright\arraybackslash}p{\dimexpr 0.2548\textwidth - 2\tabcolsep - 0.200pt\relax}>{\raggedright\arraybackslash}p{\dimexpr 0.1912\textwidth - 2\tabcolsep - 0.200pt\relax}>{\raggedright\arraybackslash}p{\dimexpr 0.2105\textwidth - 2\tabcolsep - 0.200pt\relax}@{}}
\toprule
\textbf{Decision point} & \textbf{Clinical question} & \textbf{Predictor domains admissible (cumulative)} & \textbf{Not admissible (leakage)} & \textbf{Example models} \\
\midrule
Before treatment & Start IVF? Expected chance of a baby; counselling and funding & Female age, BMI, ovarian reserve (AMH, AFC, basal FSH), infertility duration, type and cause, reproductive history; male age and semen analysis & Stimulation response; oocyte, fertilisation and embryo data & McLernon pre-treatment \cite{mclernon2016bmj}; Qiu et~al.~\cite{qiu2019jtm}; Zhang et~al.~\cite{zhang2022frontendo}; Shao et~al.~\cite{shao2026hro}; Cai et~al.~\cite{cai2024humrep}; Lantzi et~al.~\cite{lantzi2025biology}; Xing et~al.~\cite{xing2025cjrc}; stage 1 of Guo et~al.~\cite{guo2026frontendo} and Liu et~al.~\cite{liul2026fendo} \\
After ovarian stimulation & Trigger; proceed to retrieval; plan fresh transfer or freeze-all & + protocol, gonadotropin dose and duration, trigger-day hormones, endometrial thickness & Oocyte yield, fertilisation and embryo data & Peng et~al.~\cite{peng2024scirep}; stage 2 of Guo et~al.~\cite{guo2026frontendo} and Liu et~al.~\cite{liul2026fendo} \\
Before embryo transfer & Fresh transfer or freeze-all; number and choice of embryos & + oocytes retrieved, MII, 2PN, embryo development and morphology, number and stage of embryos transferred & Pregnancy test and any post-transfer event & Stage 3 of Guo et~al.~\cite{guo2026frontendo}; Banerjee et~al.~\cite{banerjee2010pnas}; Benchaib et~al.~\cite{benchaib2022rmb}; Zhu et~al.~\cite{zhu2025jor}; Wu et~al.~\cite{wu2025jtm}; single blastocyst transfer \cite{liuz2024jarg}; frozen transfer \cite{zhao2026fendo,liu2024rbejama,liang2021cmpb,baij2026cjrc}; McLernon post-treatment \cite{mclernon2016bmj} \\
After an unsuccessful complete cycle & Start another stimulation? & + full history of the previous complete cycle & Outcomes of the next cycle & Banerjee second cycle \cite{banerjee2010pnas}; Shao after the first cycle \cite{shao2026hro}; US post-treatment model \cite{mclernon2023bpr} \\
After a positive pregnancy test (prognosis only) & No treatment decision; counselling about an early pregnancy & + serum $\beta$-hCG after transfer & -- & Mourad et~al.~\cite{mourad2023fs}; Lin et~al.~\cite{lin2026fendo} \\
\bottomrule
\end{tabular}

\vspace{3pt}
{\scriptsize AFC, antral follicle count; AMH, anti-M\"{u}llerian hormone; FSH, follicle-stimulating hormone; MII, metaphase II oocytes; 2PN, two-pronuclear zygotes. ``+'' indicates that all predictors admissible at the previous decision point remain admissible.\par}
\end{table*}

\subsection{Endpoints and Units of Analysis}
``Live birth'' names a family of endpoints. Gestational thresholds range from 20 to 22 weeks in international definitions \cite{wang2023bpr} and from 22 to 28 weeks in the included studies \cite{alson2026scirep,guo2026frontendo,luo2026bmc,liu2024rbe}, and at least three studies state none \cite{peng2024scirep,wu2025jtm,zhao2026fendo}. No study modelled multiple birth separately, so a twin delivery and a singleton carry the same label, although multiple pregnancy is a harm for transfer decisions \cite{luke2014fs,wang2023bpr}. The denominator matters most: live birth per cycle started, per retrieval, per transfer or cumulatively over a complete cycle answer different questions, and models restricted to cycles that reached transfer \cite{guo2026frontendo,zhu2025jor} give probabilities conditional on reaching transfer \cite{wang2023bpr}. Eight included studies modelled fresh transfers only, sixteen cumulative live birth and eight frozen transfers only, three of them after PGT \cite{lin2026fendo,abdala2025jarg,ortiz2026jarg}; none modelled donor cycles. Cumulative endpoints lack a standard time horizon \cite{maheshwari2015humrep}, and base rates range from 22.9\% \cite{liul2026fendo} to 58.5\% \cite{yao2025natcommun}. Table~\ref{tbl:endpoints} lists the endpoint definitions and what each probability means.

\begin{table*}[!t]
\caption{Endpoint definitions and denominators in live-birth prediction, with what each probability means.}\label{tbl:endpoints}
\scriptsize
\setlength{\tabcolsep}{3.5pt}
\renewcommand{\arraystretch}{1.12}
\begin{tabular}{@{}>{\raggedright\arraybackslash}p{\dimexpr 0.1551\textwidth - 2\tabcolsep - 0.200pt\relax}>{\raggedright\arraybackslash}p{\dimexpr 0.2105\textwidth - 2\tabcolsep - 0.200pt\relax}>{\raggedright\arraybackslash}p{\dimexpr 0.2105\textwidth - 2\tabcolsep - 0.200pt\relax}>{\raggedright\arraybackslash}p{\dimexpr 0.1801\textwidth - 2\tabcolsep - 0.200pt\relax}>{\raggedright\arraybackslash}p{\dimexpr 0.2437\textwidth - 2\tabcolsep - 0.200pt\relax}@{}}
\toprule
\textbf{Endpoint} & \textbf{Numerator and threshold} & \textbf{Who is in the data (denominator)} & \textbf{Studies} & \textbf{What the predicted probability means} \\
\midrule
Clinical pregnancy (surrogate) & Pregnancy confirmed clinically; excludes later loss & Transfers & iDAScore trial \cite{illingworth2024natmed}; Sergeev and Diakova~\cite{sergeev2024scirep} & Chance of reaching clinical pregnancy; overstates the chance of live birth \\
Live birth per fresh transfer & $\geq$ 1 live infant after $\geq$ 24 weeks \cite{guo2026frontendo,liu2023bmcpc} or $\geq$ 28 weeks and alive at 1 month \cite{liu2024rbe}; multiple births count once \cite{guo2026frontendo} & Cycles that reached fresh transfer; cancelled cycles excluded & Guo et~al.~\cite{guo2026frontendo}; Zhu et~al.~\cite{zhu2025jor}; Wu et~al.~\cite{wu2025jtm}; Liu et~al.~\cite{liu2024rbe}; Benchaib (per blastocyst) \cite{benchaib2022rmb} & Chance that this transfer ends in live birth, conditional on reaching transfer \\
Live birth per frozen transfer & $\geq$ 1 live infant from the frozen transfer, often $\geq$ 28 weeks \cite{luo2026bmc} & Frozen transfers, often one per patient & Zhao et~al.~\cite{zhao2026fendo}; Luo and Wang~\cite{luo2026bmc}; Liu et~al.~\cite{liu2024rbejama}; Lin (euploid) \cite{lin2026fendo}; Abdala (euploid) \cite{abdala2025jarg}; Ortiz (mosaic) \cite{ortiz2026jarg}; Liang et~al.~\cite{liang2021cmpb}; Bai (PCOS) \cite{baij2026cjrc}; Liu (single blastocysts, 73\% frozen) \cite{liuz2024jarg} & Chance that this frozen transfer ends in live birth \\
Singleton live birth (harm-aware) & One live infant; multiple births counted as a separate outcome & Transfers or cycles & Luke et~al.~\cite{luke2014fs} (live birth and multiple birth modelled jointly) & Chance of a singleton baby; makes the harm of multiple pregnancy visible for transfer decisions \\
Live birth per treatment cycle (registry label) & ``1 if there were one or more live births as a result of this cycle'' & HFEA cycles with completed embryo transfer, fresh and subsequent frozen--thawed & Zhang et~al.~\cite{zhang2022frontendo}; Lantzi et~al.~\cite{lantzi2025biology}; Priya and Kumble~\cite{priya2026jitm} & Chance per registered cycle; women not identifiable across cycles \\
Cumulative live birth per complete cycle & $\geq$ 1 live birth from the fresh and all frozen transfers of one stimulation; a conservative variant counts untransferred embryos as no live birth \cite{fu2026fendo} & Women or retrieval cycles & Qiu et~al.~\cite{qiu2019jtm}; Fu et~al.~\cite{fu2026fendo}; Chen et~al.~\cite{chen2026fendo}; Liu et~al.~\cite{liul2026fendo}; Fatmawati et~al.~\cite{fatmawati2026sjog}; Chen et~al.~\cite{chen2022cmj}; Fraire-Zamora et~al.~\cite{frairezamora2026jarg}; Brás~de Guimarães et~al.~\cite{bras2020reprodmed}; Zhang et~al.~\cite{zhang2026jor}; Sun et~al.~\cite{sun2026clep}; Alson et~al.~\cite{alson2026scirep}; Banerjee et~al.~\cite{banerjee2010pnas}; Cai (within 2 years) \cite{cai2024humrep}; Xing et~al.~\cite{xing2025cjrc} & A woman's chance from one stimulation, including frozen transfers; time horizon not standardised \cite{maheshwari2015humrep} \\
Cumulative live birth over several complete cycles & $\geq$ 1 live birth over up to six complete cycles & Women or couples starting IVF & McLernon et~al.~\cite{mclernon2016bmj}; Shao et~al.~\cite{shao2026hro} & A woman's chance over a course of treatment \\
Live birth after a positive pregnancy test & Viable infant at $\geq$ 24 weeks \cite{mourad2023fs} or $>$ 28 weeks \cite{lin2026fendo} & Women with a positive test or a clinical pregnancy after transfer & Mourad et~al.~\cite{mourad2023fs}; Lin et~al.~\cite{lin2026fendo}; Amini et~al.~\cite{amini2021ijfs}; Bai et~al.~\cite{bai2025bmc} & Prognosis of an early pregnancy, not a prediction at a treatment decision \\
\bottomrule
\end{tabular}

\vspace{3pt}
{\scriptsize International definitions of live birth range from 20 weeks (Harbin consensus) to 22 completed weeks or $\geq$ 500 g (International Glossary) \cite{wang2023bpr}. HFEA, Human Fertilisation and Embryology Authority.\par}
\end{table*}

The unit of analysis raises three problems. If a woman's cycles fall in both training and test sets, the model is tested on patients it has seen; anonymised registry extracts make patient-grouped splits impossible \cite{zhang2022frontendo}, whereas first-cycle designs avoid the problem \cite{qiu2019jtm,guo2026frontendo,yao2025natcommun}. Embryo-level labels exist only for transferred embryos, and the outcome of a multiple-embryo transfer cannot be attributed to one embryo \cite{gao2025patterns,xi2021rbe}. The effective sample size is the number of events, and 56\% of machine-learning models across medicine have too few events per candidate predictor \cite{andaurnavarro2021bmj}.

\subsection{Data Sources and Predictor Domains}
National registries provide large samples with standardised definitions but record a minimum dataset, cannot always link cycles to women, and span changes in practice that cause calibration drift \cite{zhang2022frontendo,smith2015plosone,mclernon2023bpr}. The anonymised Human Fertilisation and Embryology Authority (HFEA) register underlies five included models \cite{goyal2020scirep,zhang2022frontendo,borji2025cmpb,jain2025informatica,priya2026jitm} and probably a sixth that does not name its source \cite{lantzi2025biology}. Single-centre records hold laboratory detail that registries lack, from 127 patients \cite{lin2026fendo} to 48,514 cycles \cite{liu2025frontendo}, but carry local coding and practice \cite{ouyang2026abme}; centre-specific models can outperform pooled registry models \cite{yao2025natcommun,yadav2026mefsj}. Predictors fall into four domains in the order they become known: baseline female and couple characteristics, male factors, stimulation, and laboratory and embryological variables. Female age dominates every family \cite{mclernon2023bpr,guo2026frontendo}, and stimulation response carried the largest incremental information in staged models \cite{guo2026frontendo}. Embryo counts are pathway-dependent: legitimate at transfer, leakage in a pre-treatment model, and not levers \cite{vangeloven2020eje}.

\section{Model Families and Pooled Performance}
\begin{figure*}[!t]
\centering
\includegraphics[width=\textwidth]{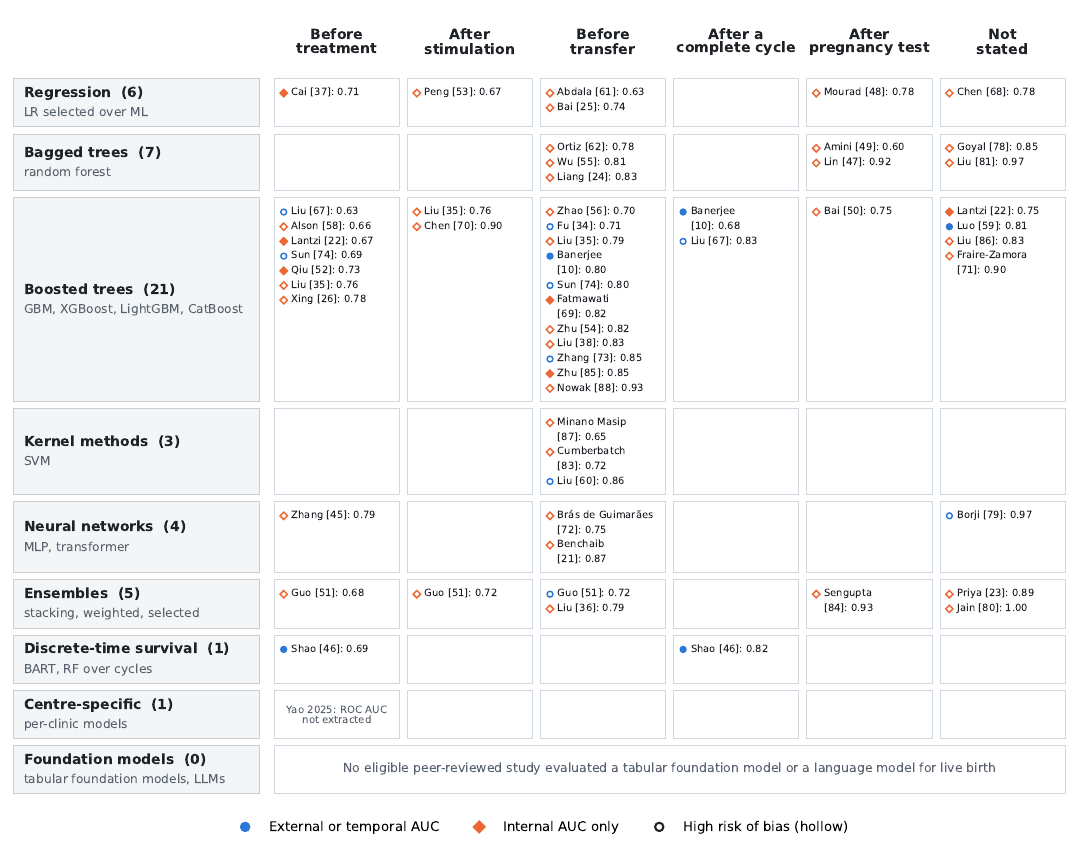}
\caption{Taxonomy of the included studies by the family of the selected model (rows) and the decision point of each estimate (columns). Each entry gives the study and its reported AUC for live birth, the external estimate where one exists; blue circles mark temporal or geographic validation, orange diamonds internal validation only, and hollow markers a high risk of bias. Numbers in parentheses are the numbers of studies in each family; numbers in square brackets are reference numbers.}\label{fig:taxonomy}
\end{figure*}

Fig.~\ref{fig:taxonomy} maps the included studies by model family, decision point and validation design. Table~\ref{tbl:studies} summarises each study with its case mix, decision point, selected model and reported AUC, and Supplementary Table D1 lists the modelling steps each reports.

\begin{table*}[!t]
\caption{Included studies of machine learning for live-birth prediction from structured data, with case mix, decision point and reported discrimination.}\label{tbl:studies}
\scriptsize
\setlength{\tabcolsep}{3.5pt}
\renewcommand{\arraystretch}{1.12}
\begin{tabular}{@{}>{\raggedright\arraybackslash}p{\dimexpr 0.0990\textwidth - 2\tabcolsep - 0.125pt\relax}>{\raggedright\arraybackslash}p{\dimexpr 0.1139\textwidth - 2\tabcolsep - 0.125pt\relax}>{\raggedright\arraybackslash}p{\dimexpr 0.1485\textwidth - 2\tabcolsep - 0.125pt\relax}>{\raggedright\arraybackslash}p{\dimexpr 0.1337\textwidth - 2\tabcolsep - 0.125pt\relax}>{\raggedright\arraybackslash}p{\dimexpr 0.0891\textwidth - 2\tabcolsep - 0.125pt\relax}>{\raggedright\arraybackslash}p{\dimexpr 0.1436\textwidth - 2\tabcolsep - 0.125pt\relax}>{\raggedright\arraybackslash}p{\dimexpr 0.1485\textwidth - 2\tabcolsep - 0.125pt\relax}>{\raggedright\arraybackslash}p{\dimexpr 0.1238\textwidth - 2\tabcolsep - 0.125pt\relax}@{}}
\toprule
\textbf{Study} & \textbf{Data and period} & \textbf{Population and endpoint} & \textbf{n; live births; female age} & \textbf{Decision point} & \textbf{Selected model (comparators)} & \textbf{AUC (95\% CI)} & \textbf{Evaluation data} \\
\midrule
Banerjee et~al.~\cite{banerjee2010pnas} & USA, 1 centre; 2003--2008 & First and second fresh autologous cycles; LB including linked frozen transfers & 1,676 development cycles; 634 + 230 validation cycles; 29\% (first cycles, development); n.a. & Before transfer; After a complete cycle & GBM boosted trees (age-only model 0.68) & Ext. 0.80 first cycle; 0.68 second cycle & temporal validation 2007--2008 \\
Qiu et~al.~\cite{qiu2019jtm} & China, 1 centre & First complete cycle; cumulative LB & 7,188 women; 2,797 (38.9\%); mean 32.7 & Before treatment & XGBoost (LR, RF, SVM) & Int. 0.73 & random 30\% test \\
Brás~de Guimarães et~al.~\cite{bras2020reprodmed} & Portugal, 1 centre; 2012--2019 & IVF/ICSI complete cycle; LB $\geq$ 24 weeks & 1,193 cycles; 387 (32.4\%); mean 34.3 & Before transfer & ANN (decision tree 0.749) & Int. 0.752 (0.725--0.775) & 15\% hold-out of rebalanced data \\
Goyal et~al.~\cite{goyal2020scirep} & UK, HFEA register & IVF cycles; LB per cycle & 141,160 after undersampling of 495,630; 50\% by design; age bands only & Not stated & RF (classical, deep and ensemble learners) & Int. 0.846 & random split of rebalanced data \\
Amini et~al.~\cite{amini2021ijfs} & Iran, 1 centre; 2011--2014 & IVF/ICSI cycles with a clinical pregnancy; successful delivery & 6,071 cycles; 4,930 (81.2\%); mean 30.9 (LB), 32.3 (no LB) & After pregnancy test & RF (LDA 0.57; LR 0.55; NB 0.53; SVM, XGBoost) & Int. 0.60 (0.55--0.64) & 30\% test, mean of 500 random splits \\
Liang et~al.~\cite{liang2021cmpb} & China, 1 centre; 2012--2015 & Frozen--thawed embryo transfers; LB $\geq$ 28 weeks per transfer & 2,189 transfer cycles; 815 (37.2\%); Not reported & Before transfer & Random forest, 100 trees, with genetic-algorithm search over modifiable factors & Int. 0.83 & 10-fold CV (0.83 $\pm$ 0.02) after predictor selection on all cycles \\
Benchaib et~al.~\cite{benchaib2022rmb} & France, 1 centre; 2013--2018 & Fresh ICSI blastocyst transfers with time-lapse culture; LB per transferred blastocyst & 654 cycles; 1,027 blastocysts; 168 deliveries in 654 cycles (25.7\%); mean 33.8 (training) & Before transfer & MLP (simple RNN 0.811, LSTM 0.861) on clinical data and time-lapse annotations & Int. 0.866 & random 30\% validation set, 294 blastocysts (196 cycles); events from the 25\% cycle-level rate; 49 of the 196 validation cycles ended in delivery, blastocyst-level events not reported \\
Chen et~al.~\cite{chen2022cmj} & China, 1 centre; 2014--2018 & Tubal or male infertility; cumulative LB $>$ 28 weeks & 3,012 patients; n.a.; n.a. & After stimulation & XGBoost (LR 0.724) & Int. 0.901 (0.890--0.912) & development data (no split described) \\
Zhang et~al.~\cite{zhang2022frontendo} & UK, HFEA register; 2005--2016 & Natural-cycle IVF; LB per cycle & 57,558 records; balanced subset 24,680; 50\% in subset; age bands only & Before treatment & ANN (LR 0.791; 12 learners) & Int. 0.794 & 10-fold CV on balanced subset \\
Liu et~al.~\cite{liu2023bmcpc} & China, 1 centre; 2019--2021 & First IVF/ICSI cycle; LB $\geq$ 24 weeks & 1,857 women; 851 (45.8\%); mean 31.9 & Before treatment; After a complete cycle & LightGBM (LR, RF, XGBoost) & Int. 0.612 (0.553--0.670) before first cycle; 0.816 (0.773--0.859) after first cycle. Ext. 0.634 (0.511--0.758) before first cycle; 0.835 (0.743--0.926) after first cycle & random 20\% validation set; temporal test year (78 women) \\
Cai et~al.~\cite{cai2024humrep} & China, 1 centre; 2013--2020 & Oocyte pick-up cycles; cumulative LB per retrieval within 2 years & 26,382 retrieval cycles; 16,673 (63.2\%); median 31.0 & Before treatment & Lasso selected (XGBoost, GLM; external McLernon, SART and Dhillon models 0.67--0.69) & Int. 0.71 (0.71--0.72) & whole cohort (development 2013--2019 and temporal set 2020) \\
Liu et~al.~\cite{liu2024rbe} & China, 1 centre & Fresh IVF/ICSI; LB $\geq$ 28 weeks, alive at 1 month & 1,405 women; 42.1\%; mean 30.6--32.2 & Before transfer & SVM (ANN 0.701) & Int. 0.854. Ext. 0.862 & random 20\% test; temporal, same centre \\
\bottomrule
\end{tabular}

\vspace{3pt}
{\scriptsize Continued on the next page.\par}
\end{table*}

\begin{table*}[!t]
\addtocounter{table}{-1}
\caption{(continued)}
\scriptsize
\setlength{\tabcolsep}{3.5pt}
\renewcommand{\arraystretch}{1.12}
\begin{tabular}{@{}>{\raggedright\arraybackslash}p{\dimexpr 0.0990\textwidth - 2\tabcolsep - 0.125pt\relax}>{\raggedright\arraybackslash}p{\dimexpr 0.1139\textwidth - 2\tabcolsep - 0.125pt\relax}>{\raggedright\arraybackslash}p{\dimexpr 0.1485\textwidth - 2\tabcolsep - 0.125pt\relax}>{\raggedright\arraybackslash}p{\dimexpr 0.1337\textwidth - 2\tabcolsep - 0.125pt\relax}>{\raggedright\arraybackslash}p{\dimexpr 0.0891\textwidth - 2\tabcolsep - 0.125pt\relax}>{\raggedright\arraybackslash}p{\dimexpr 0.1436\textwidth - 2\tabcolsep - 0.125pt\relax}>{\raggedright\arraybackslash}p{\dimexpr 0.1485\textwidth - 2\tabcolsep - 0.125pt\relax}>{\raggedright\arraybackslash}p{\dimexpr 0.1238\textwidth - 2\tabcolsep - 0.125pt\relax}@{}}
\toprule
\textbf{Study} & \textbf{Data and period} & \textbf{Population and endpoint} & \textbf{n; live births; female age} & \textbf{Decision point} & \textbf{Selected model (comparators)} & \textbf{AUC (95\% CI)} & \textbf{Evaluation data} \\
\midrule
Liu et~al.~\cite{liu2024rbejama} & China, 2 centres; 2016--2022 & Single vitrified--warmed blastocyst transfer, age $\geq$ 35; LB & 1,168 transfers; 352 (30.1\%); $\geq$ 35 & Before transfer & Stacking (19 learners) & Int. 0.791 (0.752--0.831) & random 25\% test \\
Liu et~al.~\cite{liuz2024jarg} & China, 1 centre; 2013--2019 & Single blastocyst transfers, fresh (day 5) and frozen; LB per transfer after 27 weeks & 8,383 transfers; 4,027 (48.0\%); median 30.0 & Before transfer & XGBoost with cycle features and day-3 morphology (LASSO 0.661, GLM 0.662; without day-3 morphology 0.819) & Int. 0.830 (0.822--0.839) & whole cohort, apparent (Table~3); random-split results only in Table S3 \\
Mourad et~al.~\cite{mourad2023fs} & Canada, 2 centres; 2012--2016 & Positive pregnancy test after first single transfer; LB $\geq$ 24 weeks & 1,995 patients; 1,318 (66.1\%); mean 33.3 & After pregnancy test & LR selected (classification tree, RF, NN all $\geq$ 0.70) & Int. 0.78 (0.74--0.82) & random 30\% validation \\
Peng et~al.~\cite{peng2024scirep} & China, 1 centre; 2015--2022 & IVF/ICSI cycles; LB per cycle & 11,486 couples; 3,097 (27.0\%); 33.2 (SD 5.2) & After stimulation & LR selected (RF 0.674; XGBoost, LightGBM) & Int. 0.671 (0.630--0.713) & 10-fold CV (Table~3; the abstract swaps the LR and RF values) \\
Abdala et~al.~\cite{abdala2025jarg} & UAE, 1 centre; 2017--2023 & Single euploid frozen transfer after PGT-A; LB & 1,979 transfers (1,459 couples); 1,002 (50.6\%); 51\% under 35 & Before transfer & LR selected (RF 0.606, XGBoost 0.601, SVM 0.581) & Int. 0.626 & 2,000 repeated patient-grouped 60:40 splits \\
Bai et~al.~\cite{bai2025bmc} & China, 1 centre; 2016--2022 & Fresh IVF-ET; LB among women with a clinical pregnancy & 2,031 clinical pregnancies (2,625 women); 1,711 (84.2\%); 20--40 (range) & After pregnancy test & LightGBM (LR 0.662; XGBoost 0.718; RF 0.711; 5 others) & Int. 0.746 (0.688--0.804) & 20\% test of clinical pregnancies \\
Borji et~al.~\cite{borji2025cmpb} & UK, HFEA register; 2010--2018 & IVF cycles; LB & 115,012 cycles; n.a.; age bands only & Not stated & PSO + TabTransformer (RF, DT) & Int. 0.984. Ext. 0.972 & 70/15/15 split; earlier years of the register (size not reported) \\
Cumberbatch et~al.~\cite{cumberbatch2025jsci} & Barbados, 1 centre & IVF cycles; LB & 190 cycles; n.a.; n.a. & Before transfer & Linear SVM (RBF SVM 0.63; LR 0.69) & Int. 0.72 & stratified 20\% hold-out (37 cycles) \\
Jain et~al.~\cite{jain2025informatica} & UK, HFEA register; 2010--2016 & ART cycles from the public HFEA register; LB & 2,000 cycles; n.a.; age bands only & Not stated & Stacking with XGBoost meta-learner over RF, MLP, SVM (weighted ensemble 0.994) & Int. 0.999 & stratified 20\% test \\
Lantzi et~al.~\cite{lantzi2025biology} & UK, HFEA register (source not named in the text); 2010--2018 & IVF/ICSI treatment cycles (donor-insemination cycles possibly included); LB per cycle & 665,244 cycles (stratified 50,000-row sample for training); 32,047 of 133,049 in the holdout (24.1\%); Not reported & Before treatment; Not stated & DataRobot autopilot; XGBoost with early stopping and a Keras residual network chosen (59 models screened) & Int. 0.673; 0.747 & 20\% holdout, 18 pre-cycle features (Fig.~17); 20\% holdout, 30 features including in-cycle variables (Fig.~8) \\
Liu et~al.~\cite{liu2025frontendo} & China, 1 centre & Fresh IVF cycles; LB & 48,514 cycles; n.a.; n.a. & Not stated & RF (CNN on pseudo-images 0.890; NB, DT, FNN) & Int. 0.973 & 5-fold CV and 80/20 split \\
Sengupta et~al.~\cite{sengupta2025ijcmph} & India, 1 centre; 2018--2022 & IVF/ICSI cycles; LB, stated as cumulative for the first complete cycle & 1,993 records (2,908 before exclusions); n.a.; n.a. & After pregnancy test & Bayesian-optimised voting ensemble (extra trees 0.92; RF 0.87; LightGBM 0.85; 7 learners) & Int. 0.93 & 20\% test; split ratio chosen as the best of three \\
\bottomrule
\end{tabular}

\vspace{3pt}
{\scriptsize Continued on the next page.\par}
\end{table*}

\begin{table*}[!t]
\addtocounter{table}{-1}
\caption{(continued)}
\scriptsize
\setlength{\tabcolsep}{3.5pt}
\renewcommand{\arraystretch}{1.12}
\begin{tabular}{@{}>{\raggedright\arraybackslash}p{\dimexpr 0.0990\textwidth - 2\tabcolsep - 0.125pt\relax}>{\raggedright\arraybackslash}p{\dimexpr 0.1139\textwidth - 2\tabcolsep - 0.125pt\relax}>{\raggedright\arraybackslash}p{\dimexpr 0.1485\textwidth - 2\tabcolsep - 0.125pt\relax}>{\raggedright\arraybackslash}p{\dimexpr 0.1337\textwidth - 2\tabcolsep - 0.125pt\relax}>{\raggedright\arraybackslash}p{\dimexpr 0.0891\textwidth - 2\tabcolsep - 0.125pt\relax}>{\raggedright\arraybackslash}p{\dimexpr 0.1436\textwidth - 2\tabcolsep - 0.125pt\relax}>{\raggedright\arraybackslash}p{\dimexpr 0.1485\textwidth - 2\tabcolsep - 0.125pt\relax}>{\raggedright\arraybackslash}p{\dimexpr 0.1238\textwidth - 2\tabcolsep - 0.125pt\relax}@{}}
\toprule
\textbf{Study} & \textbf{Data and period} & \textbf{Population and endpoint} & \textbf{n; live births; female age} & \textbf{Decision point} & \textbf{Selected model (comparators)} & \textbf{AUC (95\% CI)} & \textbf{Evaluation data} \\
\midrule
Wu et~al.~\cite{wu2025jtm} & China, 1 centre; 2016--2023 & Fresh cleavage-stage transfer; LB & 11,728 records; 3,971 (33.9\%); mean 31.5 (LB), 33.2 (no LB) & Before transfer & RF (XGBoost, GBM, AdaBoost, LightGBM, ANN) & Int. 0.807 & random 20\% test \\
Xing et~al.~\cite{xing2025cjrc} & China, 1 centre; 2017--2022 & IVF; cumulative LB per stimulation, fresh and frozen transfers & 1,792 cycles; 1,036 (57.8\%); mean 32.3 & Before treatment & XGBoost (RF 0.776, SVM 0.762, LR 0.728, NN 0.727) & Int. 0.775 & stratified random 30\% test (538 cycles); events from the 57.8\% rate \\
Yao et~al.~\cite{yao2025natcommun} & USA, 6 centres & First IVF cycle; LB & 4,635 patients; 56.4--58.5\% (test sets); $<$ 40 (eligibility) & Before treatment & Centre-specific ML (SART registry model) & ROC AUC not different from registry model; PR-AUC 0.75 vs 0.69 & Centre-specific out-of-time test sets \\
Zhu et~al.~\cite{zhu2025jor} & China, 1 centre; 2019--2023 & PCOS, fresh transfer; LB & 1,062 cycles; 466 (43.9\%); median 31 & Before transfer & XGBoost (DT, KNN, LightGBM, NB, RF, SVM) & Int. 0.822 (0.777--0.867) & random 30\% test \\
Zhu et~al.~\cite{zhu2025jarg} & China, 1 centre; 2018--2023 & Endometriosis (laparoscopic diagnosis), fresh transfer; LB $\geq$ 28 weeks & 1,836 patients; 739 (40.3\%); median 33 & Before transfer & XGBoost (LR 0.805; DT, KNN, LightGBM, NB, RF, SVM) & Int. 0.852 (0.821--0.884) & random 30\% test \\
Alson et~al.~\cite{alson2026scirep} & Sweden, 1 centre; 2019--2022 & First IVF/ICSI cycle, prospective; cumulative LB from the index cycle ($>$ 22 weeks) & 1,037 women; 424 (40.9\%); mean 32.0 & Before treatment & XGBoost (bagged trees post hoc) & Int. 0.66 & stratified 20\% test \\
Bai et~al.~\cite{baij2026cjrc} & China, 1 centre; 2021--2024 & PCOS, first frozen--thawed transfer; LB $\geq$ 28 weeks & 1,756 patients; 715 (40.7\%); mean 31.5 & Before transfer & Logistic regression selected (SVM 0.740, RF 0.739, XGBoost 0.732, NN 0.725) & Int. 0.745 & stratified random 30\% test (527 patients); events from the 40.7\% rate \\
Chen et~al.~\cite{chen2026fendo} & China, 1 centre; 2015--2023 & Growth-hormone co-treatment, age $<$ 40; cumulative LB & 1,782 patients; 763 (42.8\%); 34.0 (SD 3.8) & Not stated & LR selected (RF, XGBoost, LightGBM, SVM) & Int. 0.782 & 10-fold CV and bootstrap \\
Fatmawati et~al.~\cite{fatmawati2026sjog} & Indonesia, 2 centres; 2018--2023 & PCOS, first IVF/ICSI cycle; cumulative LB per initiated cycle ($\geq$ 24 weeks) & 1,245 women; 680 (54.6\%); mean 31.4 & Before transfer & XGBoost (LR 0.72; SVM 0.74; RF 0.79; GB 0.80) & Int. 0.82 (0.79--0.85) & stratified 30\% validation \\
Fraire-Zamora et~al.~\cite{frairezamora2026jarg} & Spain, multicentre; 2011--2023 & First autologous ICSI cycle; cumulative LB per completed cycle & 8,066 cycles; 3,228 (40.0\%); mean 37.3 & Not stated & GBM (LR 0.893; SVM, RF, MLP) & Int. 0.895 & nested 10-fold CV \\
Fu et~al.~\cite{fu2026fendo} & China, 1 centre; 2019--2024 & Retrieval cycles; cumulative LB & 7,773 cycles; 2,692 (34.6\%); median 34 & Before transfer & GBM (LR 0.737; AdaBoost, bagging, DT, NB) & Int. 0.762 (0.743--0.781). Ext. 0.712 (0.686--0.737) & random 30\% test; temporal test (1,607 cycles) \\
Guo et~al.~\cite{guo2026frontendo} & China, 2 centres & First fresh IVF/ICSI cycle; LB $\geq$ 24 weeks & 8,389 development + 2,058 external; 42.3--44.7\%; median 32--33 & Before treatment; After stimulation; Before transfer & Ensembles of 10 base learners & Int. 0.678 (0.649--0.708) stage 1; 0.725 (0.698--0.751) stage 2; 0.731 (0.704--0.757) stage 3. Ext. 0.719 (0.696--0.741) & locked test set; second hospital, later period \\
\bottomrule
\end{tabular}

\vspace{3pt}
{\scriptsize Continued on the next page.\par}
\end{table*}

\begin{table*}[!t]
\addtocounter{table}{-1}
\caption{(continued)}
\scriptsize
\setlength{\tabcolsep}{3.5pt}
\renewcommand{\arraystretch}{1.12}
\begin{tabular}{@{}>{\raggedright\arraybackslash}p{\dimexpr 0.0990\textwidth - 2\tabcolsep - 0.125pt\relax}>{\raggedright\arraybackslash}p{\dimexpr 0.1139\textwidth - 2\tabcolsep - 0.125pt\relax}>{\raggedright\arraybackslash}p{\dimexpr 0.1485\textwidth - 2\tabcolsep - 0.125pt\relax}>{\raggedright\arraybackslash}p{\dimexpr 0.1337\textwidth - 2\tabcolsep - 0.125pt\relax}>{\raggedright\arraybackslash}p{\dimexpr 0.0891\textwidth - 2\tabcolsep - 0.125pt\relax}>{\raggedright\arraybackslash}p{\dimexpr 0.1436\textwidth - 2\tabcolsep - 0.125pt\relax}>{\raggedright\arraybackslash}p{\dimexpr 0.1485\textwidth - 2\tabcolsep - 0.125pt\relax}>{\raggedright\arraybackslash}p{\dimexpr 0.1238\textwidth - 2\tabcolsep - 0.125pt\relax}@{}}
\toprule
\textbf{Study} & \textbf{Data and period} & \textbf{Population and endpoint} & \textbf{n; live births; female age} & \textbf{Decision point} & \textbf{Selected model (comparators)} & \textbf{AUC (95\% CI)} & \textbf{Evaluation data} \\
\midrule
Lin et~al.~\cite{lin2026fendo} & China, 1 centre; 2019--2025 & Euploid frozen transfer, recurrent pregnancy loss; LB $>$ 28 weeks & 127 patients; 79 (62.2\%); n.a. & After pregnancy test & RF (LR 0.892; seven others) & Int. 0.917 (0.778--1.000) & random 30\% test \\
Liu et~al.~\cite{liub2026fcell} & China, 1 centre; 2020--2024 & Endometrioma after sclerotherapy; cumulative LB $\geq$ 24 weeks & 194 patients; 97 (50.0\%); median 33 & Not stated & XGBoost (DT, RF, SVM) & Int. 0.830 (0.719--0.941) & random 30\% test (59 patients) \\
Liu et~al.~\cite{liul2026fendo} & China, 1 centre; 2019--2024 & Diminished ovarian reserve; cumulative LB $\geq$ 28 weeks & 1,234 patients; 282 (22.9\%); median 38 & Before treatment; After stimulation; Before transfer & CatBoost (DT, LightGBM, GBM, XGBoost, AdaBoost) & Int. 0.759 (0.704--0.810) stage 1; 0.755 (0.699--0.808) stage 2; 0.793 (0.739--0.840) stage 3 & random 30\% test \\
Luo and Wang~\cite{luo2026bmc} & China, 1 centre; 2021--2024 & Frozen--thawed transfer; LB $\geq$ 28 weeks & 3,470 development + 1,467 prospective; 41.2\% / 46.0\%; median 31 (LB), 33 (no LB) & Not stated & XGBoost (LR 0.818; RF) & Int. 0.843. Ext. 0.809 & random 30\% test; prospective temporal cohort 2024 \\
Minano~Masip et~al.~\cite{minanomasip2026medicina} & Canada, 1 centre; 2012--2019 & Single blastocyst transfers with time-lapse culture; LB & 355 transfers (259 women); 127 (35.7\%); mean 35.4 & Before transfer & SVM on clinical data (combined with a CNN on time-lapse images: 0.76) & Int. 0.65 & patient-level 15\% test \\
Nowak et~al.~\cite{nowak2026jbra} & Poland, 1 centre; 2013--2022 & IVF cycles with time-lapse annotation; LB & 375 cycles + 144 synthetic records; n.a.; n.a. & Before transfer & XGBoost (LR, SVM, NN) & Int. 0.927 & 20\% test of real plus synthetic records \\
Ortiz et~al.~\cite{ortiz2026jarg} & Spain, multicentre; 2017--2023 & Mosaic embryo transfer after PGT-A; LB & 264 transfers; 96 (36.4\%); mean 35.0 & Before transfer & RF (C5.0, SVM, GLM 0.738; 10 learners) & Int. 0.781 & random 20\% test after balancing \\
Priya and Kumble~\cite{priya2026jitm} & UK, HFEA register; 2005--2016 & Natural-cycle IVF and subsequent frozen cycles, own gametes; LB occurrence & 57,558 cycle records; n.a.; Not reported & Not stated & Ensemble of random forest and gradient boosting (RF 0.78, GB 0.77) & Int. 0.89 & test split of unreported size \\
Shao et~al.~\cite{shao2026hro} & Belgium, 1 centre; 2014--2023 & Couples starting autologous IVF; cumulative LB over up to six complete cycles ($\geq$ 24 weeks) & 1,429 development + 613 temporal couples; 848 (59.3\%); 373 (60.8\%) temporal; mean 33.1 & Before treatment; After a complete cycle & BART (elastic-net logit 0.68; RF 0.68; refitted McLernon 0.68) & Ext. 0.69 (0.65--0.73) pre-treatment; 0.82 (0.79--0.85) after first cycle & temporal validation (613 couples) \\
Sun et~al.~\cite{sun2026clep} & China, 2 centres; 2017--2023 & Women with and without ovarian endometriomas; cumulative LB from the first ART cycle ($\geq$ 28 weeks) & 1,705 development + 1,475 external women; 810 (47.5\%) development; median 32 & Before treatment; Before transfer & XGBoost (LR 0.77; DT, SVM, RF, LightGBM, NN) & Int. 0.88 (0.86--0.89) pre-ART model; 0.90 (0.88--0.92) post-retrieval model. Ext. 0.69 (0.65--0.73) pre-ART model; 0.80 (0.76--0.83) post-retrieval model & random 30\% test; second hospital (1,475) \\
Zhang et~al.~\cite{zhang2026jor} & China, 1 centre; 2019--2023 & Conventional IVF; cumulative LB per complete cycle, including frozen transfers & 1,419 cycles in 1,058 couples; 710 (50.0\%); mean 34.0 & Before transfer & GBM (LR, AdaBoost, CatBoost, ET, NN, RF, SVM) & Int. 0.827 (0.800--0.850). Ext. 0.851 (0.812--0.890) & 5-fold CV in the development cohort; temporal validation 2023 (382 cycles) \\
Zhao et~al.~\cite{zhao2026fendo} & China, 1 centre; 2021--2025 & Freeze-all, frozen transfer; LB & 990 patients; n.a.; median 34 & Before transfer & CatBoost (LR 0.680; SVM, RF, XGBoost, LightGBM) & Int. 0.704 & 10-fold CV \\
\bottomrule
\end{tabular}

\vspace{3pt}
{\scriptsize Int., internal validation (random split or cross-validation); Ext., temporal or geographic validation. Confidence intervals are shown where reported; intervals derived for the meta-analysis are given in Supplementary Table S4. n.a., not available in the sources examined. ANN, artificial neural network; ART, assisted reproductive technology; BART, Bayesian additive regression trees; CV, cross-validation; DT, decision tree; FNN, feedforward neural network; GBM, gradient-boosting machine; GLM, generalised linear model; KNN, k-nearest neighbours; LB, live birth; LR, logistic regression; ML, machine learning; NB, naive Bayes; NN, neural network; PCOS, polycystic ovary syndrome; PR-AUC, area under the precision--recall curve; PSO, particle swarm optimisation; RF, random forest; SART, Society for Assisted Reproductive Technology; SVM, support vector machine.\par}
\end{table*}

\subsection{Regression Models as the Benchmark}
Logistic and discrete-time regression set the reference. In a systematic review of 86 live-birth models, the McLernon post-treatment model had the best pooled discrimination (AUC 0.73) and the best calibration on external validation in a Chinese cohort \cite{tian2025humrep,mclernon2016bmj}. The UK registry models underestimated live-birth rates in later data, and recalibration restored performance \cite{smith2015plosone}; transported to a US centre, the Templeton model reached 0.63 \cite{banerjee2010pnas}. A regression model with a handful of pre-treatment predictors thus reaches 0.62--0.73 \cite{dhillon2016humrep,tian2025humrep}. A machine-learning model reporting 0.75 has improved on that slightly; one reporting 0.97 for the same endpoint is predicting something easier than live birth, or using information it should not have.

\subsection{Tree Ensembles and Other Classical Learners}
Random forests and gradient boosting became the default because they capture non-linearity and interactions, such as that between age and ovarian reserve \cite{gao2025patterns}, and boosted trees were the selected model in 21 of the 48 studies. Banerjee et~al.~\cite{banerjee2010pnas} reached 0.80 on a later period against 0.68 for age alone, and Qiu et~al.~\cite{qiu2019jtm} reached 0.73 with eight pre-treatment predictors in 7,188 women, equal to the McLernon benchmark. On the HFEA register, Goyal et~al.~\cite{goyal2020scirep} reported 0.846 for a random forest after removing negative records to balance the classes, for a model described as pre-treatment whose inputs include the number of embryos transferred; Zhang et~al.~\cite{zhang2022frontendo} removed every post-retrieval field and found a neural network (0.794) and logistic regression (0.791) almost indistinguishable, and Lantzi et~al.~\cite{lantzi2025biology} reached 0.673 before the cycle and 0.747 with in-cycle features on a holdout of 133,049 cycles that may include donor-insemination cycles, the largest evaluation set in this review, whereas Priya and Kumble~\cite{priya2026jitm} reported 0.89 on the same register records without stating the size of their test split. Single-centre cohorts report 0.61--0.85 \cite{liu2023bmcpc,zhu2025jor,zhu2025jarg,wu2025jtm,luo2026bmc,liang2021cmpb,xing2025cjrc}, and in several, logistic regression matched or beat the learners \cite{abdala2025jarg,baij2026cjrc,peng2024scirep}. The highest values have identifiable causes: training-set AUCs reported as test results \cite{bai2025bmc}, apparent performance on the development data \cite{chen2022cmj,liuz2024jarg}, predictors known only after treatment ends \cite{frairezamora2026jarg}, synthetic augmentation \cite{nowak2026jbra} and a 0.999 on 2,000 register records \cite{jain2025informatica}. On tabular benchmarks, trees outperform deep networks at about 10,000 samples \cite{grinsztajn2022neurips}; with a few hundred events, variance dominates and a regularised linear model is hard to beat \cite{christodoulou2019jce}.

\subsection{Deep Tabular, Foundation and Language Models}
Recent studies apply image and sequence architectures to tabular records. A convolutional network on pseudo-images of 48,514 cycles reached 0.890, while a random forest on the same data reached 0.973 \cite{liu2025frontendo}, and a TabTransformer on the HFEA register reported 0.984 \cite{borji2025cmpb}. We read near-perfect values as signs of design problems: a third to a half of infertility and early pregnancy loss remains unexplained \cite{rawlings2026humrep}, and validated models sit near 0.73 \cite{tian2025humrep}. Tabular foundation models such as TabPFN \cite{hollmann2025nature} and serialisation into language models \cite{hegselmann2023tabllm} target the small-data problem, but no peer-reviewed study has evaluated either as a live-birth predictor. The only comparison, in the authors' excluded preprint, found TabPFN (0.650) no better than logistic regression (0.667) on 38 test events \cite{shahzad2026arxiv}.

\subsection{Ensembles, Staged Models and Centre-Specific Models}
Most progress that survives external validation comes from design, not architecture. Guo et~al.~\cite{guo2026frontendo} compared ten base learners and three ensemble strategies on 8,389 fresh cycles and validated on 2,058 cycles from a second hospital: ensembles improved on the best single model by at most 0.006 AUC, whereas staging to decision points raised the AUC from 0.678 to 0.731, and the external AUC (0.719) fell only 0.012 from the internal test. Shao et~al.~\cite{shao2026hro} extended staging across cycles with discrete-time BART, reaching temporal-validation AUCs of 0.69 before treatment and 0.82 after the first cycle against a locally refitted McLernon model. Cai et~al.~\cite{cai2024humrep} found established models at 0.67--0.69 in 26,382 Chinese retrieval cycles, against 0.71 for a centre-specific Lasso model. Yao et~al.~\cite{yao2025natcommun} found that centre-specific models matched the US registry model on ROC AUC but improved precision--recall AUC and Brier score at six centres, an argument for local calibration more than for discrimination.

\subsection{Pooled Performance}
\begin{figure*}[p]
\centering
\includegraphics[width=0.97\textwidth,height=0.84\textheight,keepaspectratio]{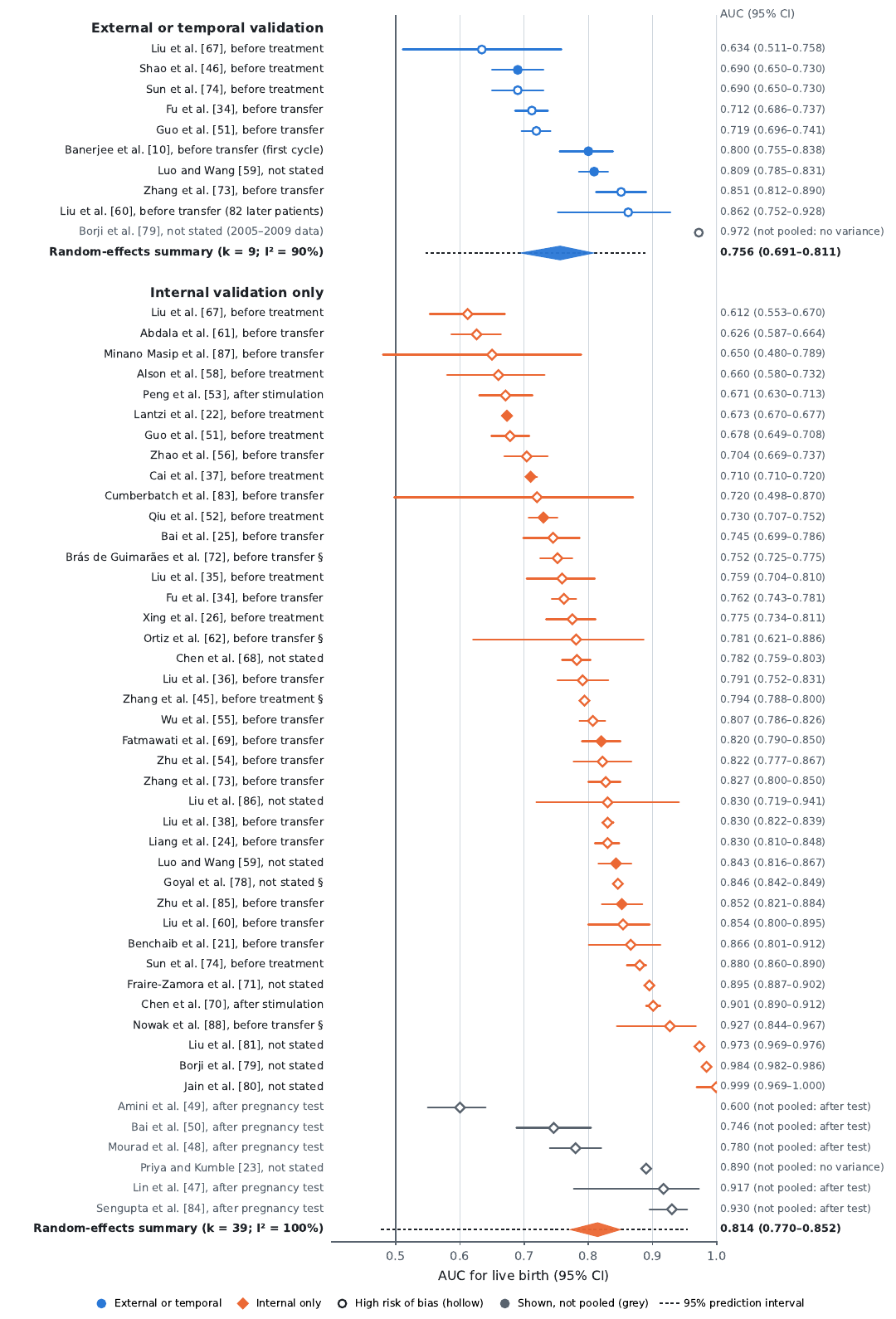}
\caption{Forest plot of AUC estimates for live birth, by validation design, with one estimate per study in each panel (Section~2.3). Diamonds are random-effects summaries on the logit scale (restricted maximum likelihood with Hartung--Knapp--Sidik--Jonkman intervals), back-transformed; dashed lines are 95\% prediction intervals. Hollow markers indicate a high risk of bias; grey estimates are shown but not pooled, because no variance could be estimated or the model used information recorded after a pregnancy test. \S\,Evaluated on rebalanced or synthetically augmented data. Intervals not reported by a study were derived as described in Section~2.3; evaluation-set sizes and event counts are given in Table~\ref{tbl:studies} and Supplementary Table S4.}\label{fig:forest}
\end{figure*}

Of the 66 estimates, 59 from 41 studies had an estimable variance (Fig.~\ref{fig:forest}). Internally validated models pooled to 0.814 (95\% CI 0.770--0.852; $k$ = 39) and models validated on a later period or another centre to 0.756 (0.691--0.811; $k$ = 9). The difference points the expected way but is not significant ($p$ = 0.11); with DerSimonian--Laird estimation it was ($p$ = 0.02), so the conclusion depends on the estimator. Heterogeneity was extreme ($I^2$ 100\% and 90\%), and the prediction intervals, 0.477--0.955 and 0.546--0.889, matter more than the means. Omitting any one study moved the estimates by at most 0.012 and 0.015. Within the seven studies with both designs at one decision point, the AUC changed by $-$0.041 ($-$0.134 to 0.033; $p$ = 0.25) and fell in four, most in the larger external sets \cite{guo2026frontendo,luo2026bmc,fu2026fendo,sun2026clep}.

Study design explains more of the spread than model family: in meta-regression, family alone explained 18\% of between-study variance, with no family significant. Studies not at high risk pooled to 0.764 (0.702--0.817; $k$ = 8), against 0.817 (0.764--0.860; $k$ = 33) at high risk ($p$ = 0.17; DerSimonian--Laird $p$ = 0.009); the one study at low risk had a temporal-validation AUC of 0.69 \cite{shao2026hro}. Pre-treatment models pooled to 0.715 (0.681--0.748; $k$ = 11), post-stimulation models to 0.779 (0.549--0.910; $k$ = 4) and pre-transfer models to 0.790 (0.759--0.818; $k$ = 23), whereas models with no stated decision point pooled to 0.910 (0.764--0.970; $k$ = 9). The moderators overlap: externally validated studies were less often at high risk of bias (6 of 9 against 27 of 32) and less often lacked a stated decision point (1 of 9 against 7 of 32). In the joint meta-regression of 41 studies, an unstated decision point raised the logit AUC by 1.12 (95\% CI 0.56--1.69; permutation $p$ = 0.001), whereas external validation ($-$0.24; $-$0.77 to 0.29) and high risk of bias (0.18; $-$0.36 to 0.72) were not significant; the moderators explained 32\% of between-study variance, and adding log standard error or model family changed nothing (decision point 1.07). An unstated decision point is the one robust marker of an inflated AUC. The internal estimate fell to 0.778 without models of unstated decision point, to 0.787 without assumed event rates and to 0.790 ($k$ = 16) with reported confidence intervals only, where the external estimate was 0.724 ($k$ = 6); cumulative and per-cycle endpoints pooled to 0.801 and 0.824. The effective-sample-size test showed no small-study effects (22 estimates; $p$ = 0.25); Egger's test ($p$ = 0.03) rested on one large register evaluation \cite{lantzi2025biology} ($p$ = 0.31 without it). The certainty of these findings is low or very low (Supplementary Table D2).

Paired comparisons, which are exploratory, point the same way. Twenty-three studies reported logistic regression on the same data, and in six it was the selected model. Across the twenty with both values, the best machine-learning model gained a median of 0.028 AUC (range $-$0.020 to +0.177), no more than 0.025 in ten, including the only temporal comparison (0.01) \cite{shao2026hro}. Pooled over the 16 studies with estimable variance, the gain was 0.33 on the logit scale (95\% CI 0.10--0.56), about 0.056 AUC (0.017--0.090) at a logistic-regression AUC of 0.75, insensitive to the assumed correlation; it was 0.055 (0.005--0.096) in the 13 studies at high risk of bias, and the three not at high risk were too few to compare (0.059; $-$0.030 to 0.125). Most larger gaps came from weak designs: tiny or rebalanced test sets \cite{cumberbatch2025jsci,ortiz2026jarg}, SMOTE \cite{sun2026clep}, post-pregnancy cohorts \cite{amini2021ijfs,bai2025bmc} and apparent performance \cite{chen2022cmj,liuz2024jarg}. The best of several tuned learners against one untuned regression favours machine learning by design; at low risk of bias, the difference across medicine is nil \cite{christodoulou2019jce}.

The validated evidence is therefore centred in the mid-0.70s with a wide spread, not at a ceiling. Higher validated values come with a reason, usually information from the treatment cycle itself \cite{banerjee2010pnas,sun2026clep,zhang2026jor,liu2023bmcpc,shao2026hro}. Discrimination is also a property of case mix: within one model, subgroup AUCs ranged from 0.628 in polycystic ovary syndrome to 0.799 in endometriosis \cite{guo2026frontendo}, so restricted cohorts yield lower AUCs from the same model. Five design features account for most high outliers: surrogate endpoints, leakage through the treatment pathway (9 of 48 studies state no decision point), random splits of one centre (only 12 of 48 validated on another centre or period), evaluation on rebalanced data \cite{goyal2020scirep,zhang2022frontendo,bras2020reprodmed,ortiz2026jarg,nowak2026jbra}, and predictors measured after transfer \cite{lin2026fendo,sengupta2025ijcmph}. No included model has prospective evidence of clinical impact. Supplementary Appendix D gives the study-by-study results and a re-check of a secondary tabulation \cite{ouyang2026abme}.

\section{Methodological Quality and Risk of Bias}
PROBAST+AI applies to prediction models built with regression or AI \cite{moons2025probastai}, unlike the diagnostic-accuracy tool used by the one previous review of AI live-birth prediction \cite{yadav2026mefsj}. It adds items that bear directly on IVF: class-imbalance corrections must be followed by recalibration and evaluation on uncorrected data, data splits must show that leakage was avoided, every development step must be repeated inside resampling, and performance must include calibration and net benefit, not apparent discrimination alone \cite{moons2025probastai}. Reporting was judged against TRIPOD+AI \cite{collins2024tripodai}.

\subsection{Participants, Predictors and Outcome}
Datasets often span a decade \cite{zhang2022frontendo,borji2025cmpb,liu2024rbejama}, over which practice and success rates change, so a temporal split measures drift directly \cite{smith2015plosone}. Missing-data handling was uneven: some studies imputed on development data only \cite{guo2026frontendo,shao2026hro}, others before splitting \cite{zhu2025jor} or did not report it \cite{zhang2022frontendo}. The most consequential predictor item asks whether predictors are available when the model is used. Goyal et~al.~\cite{goyal2020scirep} describe a pre-treatment model whose inputs include the number of embryos transferred, and the model of Borji et~al.~\cite{borji2025cmpb} selects current-cycle procedural variables and a field the paper classes as an outcome. Morphology grades vary between embryologists \cite{gao2025patterns}, so models relying on them partly learn local grading habits. Pre-processing fitted on the full dataset lets test information shape the model \cite{kapoor2023patterns}; several studies selected features, imputed or rebalanced before splitting \cite{liu2024rbe,chen2026fendo,lin2026fendo,liu2024rbejama,peng2024scirep,wu2025jtm,zhao2026fendo}. Live birth is recorded objectively, but definitions, follow-up and the horizon of cumulative endpoints vary \cite{maheshwari2015humrep}.

\subsection{Analysis}
The analysis domain is where machine-learning models most often fail \cite{andaurnavarro2021bmj}. Development samples should be sized with formal criteria \cite{riley2020bmj}; some included studies compared several learners on development sets of 90--140 patients \cite{liub2026fcell,lin2026fendo}. Evaluation needs calibration, the Brier score and net benefit:
\begin{align}
\operatorname{logit}\Pr(Y_i=1)&=\alpha+\beta\,\operatorname{logit}(\hat p_i),\label{eq:cal}\\
\mathrm{BS}&=\frac{1}{n}\sum_{i=1}^{n}(\hat p_i-y_i)^2,\label{eq:brier}\\
\mathrm{NB}(p_t)&=\frac{\mathrm{TP}(p_t)}{n}-\frac{\mathrm{FP}(p_t)}{n}\cdot\frac{p_t}{1-p_t},\label{eq:nb}
\end{align}
where the slope $\beta$ should be 1 and the intercept $\alpha$, estimated with $\beta$ fixed at 1, should be 0 \cite{vancalster2019bmcmed}; the Brier score combines discrimination and calibration; and net benefit at threshold $p_t$ is plotted in decision-curve analysis \cite{vickers2006mdm}. External validation needs at least 100 and preferably 200 events \cite{collins2016statmed}, but several test sets held fewer than 60 patients \cite{lin2026fendo,liub2026fcell,cumberbatch2025jsci}, with intervals that span most competitors. Repeated cross-validation or bootstrapping is preferred to a single split \cite{steyerberg2016jce}; one study described no split \cite{chen2022cmj}, and five reused test or validation data for model selection \cite{liu2024rbe,sengupta2025ijcmph,liul2026fendo,fu2026fendo,zhang2026jor}.

All four classes of leakage in the taxonomy of Kapoor and Narayanan \cite{kapoor2023patterns} occur: non-independent splits \cite{zhang2022frontendo}, pre-processing before the split \cite{zhu2025jor,liu2024rbe,bras2020reprodmed,ortiz2026jarg}, illegitimate features \cite{goyal2020scirep,borji2025cmpb,frairezamora2026jarg} and temporal leakage. Class imbalance is modest in IVF (live birth in 23--62\% of most cohorts), yet several studies rebalanced, which degrades calibration without improving discrimination \cite{vandengoorbergh2022jamia}, and evaluated on rebalanced or synthetic data \cite{zhang2022frontendo,goyal2020scirep,bras2020reprodmed,ortiz2026jarg,nowak2026jbra}; four applied SMOTE without recalibration \cite{borji2025cmpb,liu2024rbejama,jain2025informatica,sun2026clep}. Across medicine, 87\% of supervised machine-learning analyses were at high risk of bias, most often in the analysis domain \cite{andaurnavarro2021bmj}, and comparisons at high risk favoured machine learning over logistic regression by 0.34 on the logit-AUC scale, against 0.00 at low risk \cite{christodoulou2019jce}. Fair comparison needs equal tuning budgets, as in the shared Optuna budget of Guo et~al.~\cite{guo2026frontendo}, whereas some studies reported tuning only for the proposed model \cite{liu2025frontendo} or reported no tuning \cite{zhang2022frontendo}; only one study refitted the reference regression model on its own data \cite{shao2026hro}. Of the 48 studies, 28 reported calibration in some form and 18 quantitatively, but only four reported the calibration slope \cite{shao2026hro,abdala2025jarg,fatmawati2026sjog,cai2024humrep}, and 12 reported decision-curve analysis.

\subsection{Applicability, Reporting and Overall Appraisal}
Subgroup AUCs partly reflect a narrower range of risk, so calibration within subgroups is the better test of whether a model serves each group \cite{vancalster2019bmcmed}; no study reported it. Nine studies did not state the decision point, one released code \cite{wu2025jtm}, none reported patient involvement, and several gave inconsistent figures between sections \cite{liu2025frontendo,zhang2022frontendo,wu2025jtm,borji2025cmpb}. After adjudication one study was at low overall risk of bias \cite{shao2026hro}, 7 at unclear and 40 at high risk (Supplementary Table E1); 37 of the high ratings involved the analysis domain. The second appraisals were stricter than the original ratings: raw agreement was 0.55 and 0.65 over the four domains and 0.59 and 0.68 for the overall rating, where kappa was only 0.09 and 0.11 because nearly every study was rated high (AC1 0.47 and 0.62). We changed 42 of 192 domain judgements and 11 overall ratings, 10 towards higher risk, among them Qiu, Fatmawati and Guo et~al.\ from low \cite{qiu2019jtm,fatmawati2026sjog,guo2026frontendo} (Supplementary Table E2). Two models combine leakage control, independent units, adequate test events, external validation and calibration in their analysis: the staged ensemble of Guo et~al.~\cite{guo2026frontendo}, whose baseline stage was nevertheless developed only on cycles that reached transfer, and the discrete-time model of Shao et~al.~\cite{shao2026hro}. The models with the highest reported discrimination carry the most serious concerns (Supplementary Appendix E).

\section{Interpretability, Fairness and Clinical Translation}
\subsection{What Explanations Show, and What They Do Not}
Studies explain models with native importance, post hoc attribution such as SHAP \cite{lundberg2017shap}, or attention weights. All such methods share three limits. Explanations describe the model, not the biology, and different models fitted to the same data rank predictors differently \cite{shahzad2026arxiv}. And association is not intervention: doses and protocols are chosen in response to patient characteristics, so an optimal dose read from SHAP values \cite{guo2026frontendo} is a hypothesis for trials, not a treatment rule \cite{vangeloven2020eje}. Because the evidence that complex models beat well-specified regression comes mostly from studies at high risk of bias, the burden of proof lies with the complex model \cite{rudin2019natmachintell,christodoulou2019jce}.

\subsection{Fairness and Equity}
TRIPOD+AI asks for performance in key subgroups and a discussion of fairness \cite{collins2024tripodai}. Under-prediction has a direct cost: the US registry model gave lower probabilities than centre-specific models \cite{yao2025natcommun}, and systematically low estimates can discourage treatment or block funding. Representation is narrow. Twenty-six of the 48 studies come from China and six from the UK; the rest come from 13 countries, none in Africa. Models trained mainly on one population may mis-predict for others \cite{gao2025patterns}, and compact models that can be retrained locally make recalibration and audit practical in low-resource settings.

\subsection{From Prediction to Clinical Use}
A probability changes a decision only through a threshold that weighs false reassurance against false discouragement \cite{vickers2006mdm}; twelve studies reported decision-curve analysis, none with pre-specified thresholds. Transported models drift and need local recalibration and monitoring \cite{smith2015plosone,yao2025natcommun}. Impact evidence is missing: no structured-data live-birth model has been tested for impact on decisions or outcomes, and the one large trial of an AI embryo-ranking tool did not demonstrate non-inferiority \cite{illingworth2024natmed,leushuis2009hru}. Clinical deployment carries regulatory obligations in the European Union \cite{eumdr2017,euaiact2024,euomnibus2026}, the United States \cite{imdrf2014samd,fda2024pccp} and China \cite{nmpa2021classification,cmde2022registration}; all converge on data governance, documentation, human oversight, change control and post-market monitoring, which compact, locally auditable models meet more easily \cite{cohen2025rbmo}. Table~\ref{tbl:translation} maps the path from development to routine use.

\begin{table*}[!t]
\caption{Stages from model development to routine use, and the status of structured-data live-birth models.}\label{tbl:translation}
\scriptsize
\setlength{\tabcolsep}{3.5pt}
\renewcommand{\arraystretch}{1.12}
\begin{tabular}{@{}>{\raggedright\arraybackslash}p{\dimexpr 0.1662\textwidth - 2\tabcolsep - 0.250pt\relax}>{\raggedright\arraybackslash}p{\dimexpr 0.2327\textwidth - 2\tabcolsep - 0.250pt\relax}>{\raggedright\arraybackslash}p{\dimexpr 0.2799\textwidth - 2\tabcolsep - 0.250pt\relax}>{\raggedright\arraybackslash}p{\dimexpr 0.3213\textwidth - 2\tabcolsep - 0.250pt\relax}@{}}
\toprule
\textbf{Stage} & \textbf{Question answered} & \textbf{Evidence required} & \textbf{Status in the literature reviewed} \\
\midrule
1. Development & Does the model discriminate and calibrate in its own data? & Internal validation with optimism correction; calibration; TRIPOD+AI reporting \cite{collins2024tripodai} & Abundant; mostly single-centre, most at high risk of bias (Sections~4 and 5) \\
2. External validation & Does performance hold in other centres and later periods? & Temporal and geographic validation with $\geq$ 100--200 events; calibration \cite{collins2016statmed} & Few: McLernon models \cite{ratna2023humrep,tian2025humrep,cai2024humrep}; Banerjee et~al.~\cite{banerjee2010pnas}; Guo et~al.~\cite{guo2026frontendo}; Fu et~al.~\cite{fu2026fendo}; Luo and Wang~\cite{luo2026bmc}; Shao et~al.~\cite{shao2026hro}; Sun et~al.~\cite{sun2026clep}; Zhang et~al.~\cite{zhang2026jor}; centre-specific models \cite{yao2025natcommun} \\
3. Local updating & Is the model calibrated for this clinic now? & Recalibration or refitting on recent local data; monitoring for drift & Recalibration improved IVFpredict \cite{smith2015plosone}; Platt scaling \cite{guo2026frontendo,liul2026fendo}; McLernon model refitted locally \cite{shao2026hro}; centre-specific modelling \cite{yao2025natcommun} \\
4. Clinical utility & Would using the model improve decisions? & Decision-curve analysis at pre-specified thresholds \cite{vickers2006mdm} & Reported in 12 of the 48 included studies; thresholds not pre-specified \\
5. Impact & Does using the model change decisions and outcomes? & Prospective comparative study, ideally randomised & None identified; nearest are the iDAScore trial \cite{illingworth2024natmed} and the IDoser protocol \cite{gao2025patterns} \\
6. Implementation and monitoring & Does the model stay safe and calibrated in routine use? & Post-market monitoring, change control and regulatory conformity \cite{euaiact2024,fda2024pccp,cmde2022registration} & Not reported; EU high-risk obligations for AI in medical devices apply from 2 August 2028 \cite{euomnibus2026} \\
\bottomrule
\end{tabular}

\vspace{3pt}
{\scriptsize Stages 1--3 concern the validity of predictions; stages 4--6 concern their value and safety in care.\par}
\end{table*}

\section{Emerging Directions and the Methods They Require}
Table~\ref{tbl:emerging} summarises the promise, evidence, enabling methods and risks of the directions named by Gao et~al.~\cite{gao2025patterns}.

\begin{table*}[!t]
\caption{Emerging directions for live-birth prediction: promise, current evidence, enabling methods, main risk, and what a convincing study would show.}\label{tbl:emerging}
\scriptsize
\setlength{\tabcolsep}{3.5pt}
\renewcommand{\arraystretch}{1.12}
\begin{tabular}{@{}>{\raggedright\arraybackslash}p{\dimexpr 0.1437\textwidth - 2\tabcolsep - 0.167pt\relax}>{\raggedright\arraybackslash}p{\dimexpr 0.1609\textwidth - 2\tabcolsep - 0.167pt\relax}>{\raggedright\arraybackslash}p{\dimexpr 0.1954\textwidth - 2\tabcolsep - 0.167pt\relax}>{\raggedright\arraybackslash}p{\dimexpr 0.1724\textwidth - 2\tabcolsep - 0.167pt\relax}>{\raggedright\arraybackslash}p{\dimexpr 0.1552\textwidth - 2\tabcolsep - 0.167pt\relax}>{\raggedright\arraybackslash}p{\dimexpr 0.1724\textwidth - 2\tabcolsep - 0.167pt\relax}@{}}
\toprule
\textbf{Direction} & \textbf{Promise} & \textbf{Evidence for live birth so far} & \textbf{Enabling methods} & \textbf{Main risk} & \textbf{A convincing study would show} \\
\midrule
Multimodal fusion (images + clinical data) & Adds embryo and endometrial information that records lack & Fusion beat images alone (0.77 vs 0.70) \cite{liu2023elife}; fusion beat tabular data alone by 0.033 \cite{ouyang2025defusion} & Decoupled or late fusion; image foundation models \cite{rajendran2025femi} & Comparison against images alone; equipment-specific inputs & Gain over a tuned tabular model, externally validated, with calibration \\
Tabular foundation models & Strong small-sample performance without tuning & Strong on public benchmarks \cite{hollmann2025nature}; one live-birth cohort, in a preprint \cite{shahzad2026arxiv} & In-context learning on synthetic priors & Prior may not match clinical data & Head-to-head comparison across cohorts, with calibration \\
Language models & Use of free text and prior knowledge; automated pipelines & Few-shot strength on public tasks \cite{hegselmann2023tabllm}; pregnancy only \cite{cao2025aogs} & Serialisation, fine-tuning, information extraction & Loss of numerical structure; leakage hidden in automated pipelines \cite{fu2025ejmo} & Live-birth endpoint, admissible predictors, comparison with logistic regression \\
Federated learning & Multicentre training without moving records & Clinical-pregnancy AUC 0.57--0.72 across four hospitals \cite{lee2024jarg} & FedAvg \cite{mcmahan2017aistats}; FedProx \cite{li2020mlsys}; secure aggregation \cite{bonawitz2017ccs}; differential privacy \cite{abadi2016ccs} & Between-centre heterogeneity; leakage through updates & Live-birth model compared with central and local training, with per-centre calibration \\
Dynamic prediction & Updates as treatment progresses & Staged 0.678 $\rightarrow$ 0.731 \cite{guo2026frontendo}; discrete-time BART over six cycles \cite{shao2026hro}; adaptive per-transfer models \cite{grzegorczyk2022jarg} & Discrete-time and multi-state models \cite{putter2007sim}; landmarking \cite{vanhouwelingen2007sjs}; joint models \cite{rizopoulos2011biometrics} & Leakage across stages; informative discontinuation & Stage-isolated predictors; cumulative endpoint over all transfers \\
Prediction under intervention & Chance of live birth under each treatment option & None identified & Predictimands \cite{vangeloven2020eje}; meta-learners \cite{kunzel2019pnas}; causal forests \cite{wager2018jasa} & Confounding by indication & Randomised-trial data or stated causal assumptions \\
Uncertainty for counselling & Individual reliability of a prediction & None identified & Recalibration \cite{platt1999,zadrozny2002kdd}; conformal prediction \cite{angelopoulos2023ftml} & Uninformative sets for many patients & Coverage and set size reported on external data \\
Digital twins & Simulate treatment scenarios before acting & Conceptual \cite{gao2025patterns,olawade2026jogoh} & All of the above & Unvalidated simulation assumptions & Prospective comparison of simulated with observed outcomes \\
\bottomrule
\end{tabular}

\vspace{3pt}
{\scriptsize AUC values are as reported by the cited studies; endpoints differ between rows and are named where they are not live birth.\par}
\end{table*}

\subsection{Multimodal Fusion}
Images add embryo and endometrial information that records lack. Fused models reached 0.77 for live birth against 0.70 for images alone \cite{liu2023elife}, IVFormer combined pre-trained embryo image and video representations with clinical data to reach 0.854 \cite{wang2024patterns}, and hybrid video--tabular models reached 0.73 for fetal heartbeat across 14 centres \cite{duval2023humrep}, but the relevant comparison is against a strong tabular model: in DeFusion, images added 0.033 AUC to clinical data, whereas clinical data added 0.129 to images \cite{ouyang2025defusion}. Embryo foundation models such as FEMI \cite{rajendran2025femi} have not been evaluated on live birth.

\subsection{Foundation Models and Language Models}
The informative study of tabular foundation models is a head-to-head comparison across cohorts with equal tuning and calibration \cite{hollmann2025nature}. Language models that write pipelines \cite{cao2025aogs} can hide leakage: one automated pipeline's top predictor was $\beta$-hCG seven days after transfer \cite{fu2025ejmo}. Their safer uses are extracting structured fields from free-text records and communicating validated predictions.

\subsection{Federated Learning, Privacy and Shared Benchmarks}
Federated averaging \cite{mcmahan2017aistats} and FedProx \cite{li2020mlsys}, with secure aggregation \cite{bonawitz2017ccs} and differential privacy \cite{abadi2016ccs}, train across centres without moving records. Federated boosted trees across four hospitals reached 0.568--0.720 for clinical pregnancy \cite{lee2024jarg}, a spread that shows federation does not remove case-mix differences. Structured live-birth research has no public benchmark beyond the HFEA register; Table~\ref{tbl:benchmark} specifies one with fixed decision points, endpoints and denominators, patient-grouped temporal and geographic splits, equal-budget baselines and calibration and net-benefit metrics.

\begin{table*}[!t]
\caption{Specification of a shared benchmark for live-birth prediction from structured IVF/ICSI data.}\label{tbl:benchmark}
\scriptsize
\setlength{\tabcolsep}{3.5pt}
\renewcommand{\arraystretch}{1.12}
\begin{tabular}{@{}>{\raggedright\arraybackslash}p{\dimexpr 0.1994\textwidth - 2\tabcolsep - 0.500pt\relax}>{\raggedright\arraybackslash}p{\dimexpr 0.8006\textwidth - 2\tabcolsep - 0.500pt\relax}@{}}
\toprule
\textbf{Element} & \textbf{Specification} \\
\midrule
Data & The anonymised HFEA register as a public seed; contributed single-centre and multicentre datasets under controlled access, or evaluated in place by federated evaluation \\
Schema & A minimal data dictionary of the four predictor domains (Table~2), with each variable time-stamped to the decision point at which it becomes known, and the cycle, transfer and woman identifiers needed for grouping \\
Tasks & T1 before treatment: cumulative live birth per complete cycle. T2 after stimulation: live birth per fresh transfer. T3 before transfer: live birth per transfer, fresh and frozen. T4 after an unsuccessful cycle: live birth in the next complete cycle \\
Endpoints & Live birth with a stated gestational threshold (International Glossary definition by default); multiple births counted once; singleton live birth at term as a secondary endpoint for T2 and T3 \\
Splits & Patient-grouped; a temporal hold-out of the most recent two years; a geographic hold-out of at least one centre; no tuning on hold-out data \\
Baselines & Penalised logistic regression with splines; the McLernon models for T1 and T4; gradient-boosted trees; a tabular foundation model; equal tuning budgets \\
Metrics & AUC with 95\% CI and paired differences; calibration intercept and slope with curves; Brier score; net benefit at pre-specified thresholds; all reported by age band, principal diagnosis and centre \\
Reporting and governance & TRIPOD+AI checklist and model card; code and trained models released; hidden test labels held by an independent steward; versioned releases \\
\bottomrule
\end{tabular}

\vspace{3pt}
{\scriptsize The specification follows Sections~3 and 5 and TRIPOD+AI \cite{collins2024tripodai}.\par}
\end{table*}

\subsection{Dynamic, Survival and Counterfactual Prediction}
Discrete-time and multi-state models \cite{mclernon2016bmj,shao2026hro,putter2007sim}, landmarking \cite{vanhouwelingen2007sjs} and joint models \cite{rizopoulos2011biometrics} can update predictions as treatment proceeds, treating discontinuation as a competing event. The decisions in Section~3.1 are choices between treatments, and observational models predict only under the treatment received; predictimands \cite{vangeloven2020eje}, meta-learners \cite{kunzel2019pnas} and causal forests \cite{wager2018jasa}, fitted to data from randomised trials of freeze-all and fresh transfer \cite{stormlund2020bmj,wei2025bmj}, could estimate each patient's benefit from each option. No such model exists for live birth.

\subsection{Recalibration and Uncertainty for Individual Counselling}
Updating proceeds from recalibration in the large through logistic recalibration to refitting \cite{steyerberg2019book}; Platt scaling \cite{platt1999} and isotonic regression \cite{zadrozny2002kdd} map machine-learning scores to probabilities, and two included studies used Platt scaling \cite{guo2026frontendo,liul2026fendo}. Conformal prediction \cite{angelopoulos2023ftml} would show, for each woman, whether routine data can separate her outcomes at all. No included study quantified individual uncertainty.

\section{Recommendations for Future Studies}
\subsection{Developing and Validating Models}
A new model is justified only when it serves a decision point existing models do not, uses information they lack, or targets a population in which they are miscalibrated, and it must show added value over them on the same data \cite{tian2025humrep}. Table~\ref{tbl:checklist} lists sixteen items mapped to TRIPOD+AI. Three are missed most often: the decision point belongs in the title or abstract, with one model per stage \cite{guo2026frontendo}; every model needs equally tuned baselines, at minimum penalised logistic regression with splines and an established model, with paired differences \cite{christodoulou2019jce}; and calibration and net benefit at pre-specified thresholds must be reported \cite{vancalster2019bmcmed,vickers2006mdm}. External validation of the best existing models \cite{ratna2023humrep,tian2025humrep,cai2024humrep,guo2026frontendo}, sized by formal criteria \cite{riley2021statmed} and reported per centre \cite{lee2024jarg}, is worth more than another single-centre model.

\begin{table*}[!t]
\caption{Checklist for studies developing or validating live-birth prediction models from structured IVF/ICSI data, mapped to TRIPOD+AI items.}\label{tbl:checklist}
\scriptsize
\setlength{\tabcolsep}{3.5pt}
\renewcommand{\arraystretch}{1.12}
\begin{tabular}{@{}>{\raggedright\arraybackslash}p{\dimexpr 0.1163\textwidth - 2\tabcolsep - 0.200pt\relax}>{\raggedright\arraybackslash}p{\dimexpr 0.5125\textwidth - 2\tabcolsep - 0.200pt\relax}>{\raggedright\arraybackslash}p{\dimexpr 0.0997\textwidth - 2\tabcolsep - 0.200pt\relax}>{\raggedright\arraybackslash}p{\dimexpr 0.1053\textwidth - 2\tabcolsep - 0.200pt\relax}>{\raggedright\arraybackslash}p{\dimexpr 0.1662\textwidth - 2\tabcolsep - 0.200pt\relax}@{}}
\toprule
\textbf{Stage} & \textbf{Recommendation} & \textbf{Section} & \textbf{TRIPOD+AI item} & \textbf{Key reference} \\
\midrule
Question & 1. State the decision point and intended user; admit only predictors routinely available at that point. & 3.1, 5.1 & 3, 9 & \cite{moons2025probastai,guo2026frontendo} \\
Question & 2. Define the endpoint fully: gestational threshold, numerator, denominator; report the composition of the non-live-birth group and, for transfer decisions, singleton live birth. & 3.2 & 8, 20 & \cite{wang2023bpr,maheshwari2015humrep} \\
Data & 3. Use one cycle per woman, or group data splits by patient; report how repeated cycles were handled. & 3.2, 5.2 & 6, 12 & \cite{kapoor2023patterns} \\
Data & 4. Report the data-collection period and prefer a temporal test set. & 5.1 & 5, 12 & \cite{smith2015plosone,steyerberg2016jce} \\
Data & 5. Report missingness per variable; impute with methods fitted on training data only. & 5.1 & 7, 11 & \cite{andaurnavarro2021bmj} \\
Sample size & 6. Justify the development sample size with formal criteria; size external validation for precise calibration (at least 100 events, usually more). & 5.2, 8.1 & 10 & \cite{riley2020bmj,riley2021statmed,collins2016statmed} \\
Modelling & 7. Fit all pre-processing and feature selection within training folds. & 5.1, 5.2 & 7, 12 & \cite{kapoor2023patterns} \\
Modelling & 8. Avoid class-imbalance corrections; if used, recalibrate and evaluate on uncorrected data. & 5.2 & 13 & \cite{vandengoorbergh2022jamia,moons2025probastai} \\
Modelling & 9. Benchmark against penalised logistic regression with splines and an established model, with equal tuning budgets. & 4.2, 5.2, 8.1 & 12 & \cite{christodoulou2019jce,tian2025humrep} \\
Evaluation & 10. Report discrimination with confidence intervals and paired comparisons; calibration intercept, slope and curve; Brier score; net benefit at pre-specified thresholds. & 5.2, 6.3 & 12, 23 & \cite{vancalster2019bmcmed,vickers2006mdm} \\
Evaluation & 11. Report calibration and discrimination by age band and principal diagnosis. & 5.3, 6.2 & 14, 23 & \cite{guo2026frontendo,collins2024tripodai} \\
Evaluation & 12. Validate temporally and geographically; report case-mix differences and per-centre results. & 4.5, 8.1 & 16, 23 & \cite{steyerberg2016jce,guo2026frontendo} \\
Interpretation & 13. Report feature importance as a consensus across models; do not read SHAP values or attention weights as causal effects. & 6.1 & 25 & \cite{jain2019attention,vangeloven2020eje} \\
Reporting & 14. Follow TRIPOD+AI; share code and the full model; state the conditions for data access. & 5.3 & 18, 22 & \cite{collins2024tripodai} \\
Reporting & 15. Involve patients in choosing the endpoint and in how probabilities are communicated. & 6.2 & 19 & \cite{collins2024tripodai} \\
Translation & 16. Plan local recalibration, monitoring for drift, change control and regulatory requirements before deployment. & 6.3, 7.5 & 24, 27 & \cite{smith2015plosone,euaiact2024,fda2024pccp} \\
\bottomrule
\end{tabular}

\vspace{3pt}
{\scriptsize The checklist was derived by the authors from this review, not by a consensus process. Items 1--2 and 13 are specific to the IVF setting; the remaining items apply TRIPOD+AI and PROBAST+AI to live-birth prediction. TRIPOD+AI items are numbered as in the published checklist \cite{collins2024tripodai}.\par}
\end{table*}

\subsection{Evidence Synthesis, Peer Review and Funding}
Reviewers should appraise with PROBAST+AI, stratify by decision point, endpoint and denominator, keep development estimates apart from external ones, and pool the c-statistic on the logit scale with REML, Hartung--Knapp--Sidik--Jonkman intervals and prediction intervals \cite{moons2025probastai,snell2018smmr,debray2017bmj}. Peer reviewers should ask for an explanation when a structured-data model reports an AUC above about 0.85 for live birth without external validation, states no decision point, includes post-transfer or outcome-derived variables in a model presented as pre-transfer, rebalances without recalibration \cite{vandengoorbergh2022jamia}, leaves fewer than 100 test events \cite{collins2016statmed}, ranks models by accuracy or F1 at a 0.5 threshold, or reports training-set performance in the abstract \cite{bai2025bmc}. Funders should shift support from new model development towards external validation, impact studies and shared infrastructure.

\subsection{Prioritised Research Agenda}
The priorities are, first, external validation and recalibration of the best existing models in new countries and recent cohorts \cite{tian2025humrep,guo2026frontendo,yao2025natcommun}; second, a public benchmark (Table~\ref{tbl:benchmark}); and third, a randomised impact study of a validated pre-treatment counselling model. Supplementary Appendix G adds four methodological lines: prediction under alternative treatments, multi-state models, multimodal models against tuned tabular baselines, and federated models with per-centre calibration.

\section{Limitations of This Review}
Screening was done by a single assessor, so eligible studies may have been missed. Risk of bias was re-appraised independently by two co-authors, who read the full texts but used AI to draft and check their ratings, so agreement reflects AI-assisted appraisal rather than unaided experts. The search did not cover Embase, IEEE Xplore or the Chinese databases CNKI and Wanfang, although more than half the studies are Chinese, and Scopus added six studies that OpenAlex had missed, so other studies may have been missed. Several full texts did not report the data used for imputation or feature selection, which leaves some ratings unclear. The meta-analysis pools heterogeneous estimands: each study contributed its selected model, internal estimates mix random splits with cross-validation, some standard errors were derived, and selective reporting probably inflates internal estimates further; the protocol was not registered, several analyses are exploratory, and certainty is low to very low. Three of the authors have developed a live-birth model \cite{shahzad2026arxiv}, excluded from the evidence, and readers should weigh the discussion of compact models with this in mind. Thirty-eight of the 48 studies were published in 2024 or later, so the conclusions should be revisited as new external validations appear.

\section{Conclusions}
Live-birth prediction from structured IVF data has a large literature but a thin evidence base. A model's output means something only in relation to its decision point, endpoint and denominator. Across 48 studies, externally validated models pooled to an AUC of 0.756 (prediction interval 0.546--0.889) and studies not at high risk of bias to 0.764, against 0.814 for internal validation and 0.910 for models without a stated decision point; these contrasts are consistent in direction but not significant, and certainty is low. In a joint meta-regression, only an unstated decision point remained associated with a higher AUC. Against logistic regression on the same data, the best machine-learning model gained a pooled 0.056 AUC (95\% CI 0.017--0.090), 13 of whose 16 comparisons were at high risk of bias. The highest values coincide with surrogate endpoints, leakage, rebalanced evaluation, non-independent splits and small samples. The gains that survive validation came from staging predictions to decision points and calibrating locally, not from deeper architectures, and no model has evidence of clinical impact. For clinicians, a well-calibrated, transparent model built for a stated decision point is worth more than a higher AUC; for developers, the bar is added value over validated existing models on the same data.

\section*{Conflict of Interest}
The authors declare no competing financial interests. A.R.S., M.Z. and J.D. are authors of a preprint on live-birth prediction discussed in this review \cite{shahzad2026arxiv}; under the review's preprint rule it was excluded from the evidence tables, the appraisal and the meta-analysis.

\section*{Use of Generative AI}
The authors used Claude (Anthropic) to support literature searching, location of reported figures, drafting and language editing; M.Z.K. and U.A. read the full texts and used it to draft and check their second appraisals. Every included study and extracted value was checked by the first author against the original article, and the authors take full responsibility for the content.

\section*{Funding and Data Availability}
This research received no specific funding. All data and code are in Supplementary Files S1--S10 and at \url{https://github.com/abdul-raheem-shahzad/ml-ivf-live-birth-review}. No new patient data were generated.

\bibliographystyle{IEEEtran}
\bibliography{refs}

\end{document}